\documentclass{article}

    \PassOptionsToPackage{numbers,sort&compress}{natbib}

    \usepackage[preprint]{neurips_2025}

\usepackage[utf8]{inputenc} 
\usepackage[T1]{fontenc}    
\usepackage{hyperref}       
\usepackage{url}            
\usepackage{booktabs}       
\usepackage{amsfonts}       
\usepackage{nicefrac}       
\usepackage{microtype}      
\usepackage{xcolor}         
\usepackage{graphicx}
\usepackage{amsmath}
\usepackage{bm}
\usepackage{algorithm}
\usepackage{algpseudocode}
\usepackage{wrapfig}
\usepackage{caption}
\usepackage{xspace}
\usepackage{hyperref}

\title{\textit{Half-Physics}: Enabling Kinematic 3D Human Model\\with Physical Interactions}

\newcommand{\hp}{\textit{half physics}}%
\newcommand{\ie}{\textit{i.e.}}
\newcommand{\eg}{\textit{e.g.}}
\newcommand{\etal}{\textit{et al.}}

\newcommand{\LightComment}[1]{\Comment{\textcolor{blue}{#1}}}

\author{%
 Li Siyao$^{1,2}$\thanks{work done during internship at MPI-IS}, \ \ Yao Feng$^3$, \ \ Omid Taheri$^1$, \ \ Chen Change Loy$^2$, \ \ Michael J. Black$^1$ \\
  $^1$Max Planck Institute for Intelligent Systems, T\"ubingen\\
  $^2$S-Lab, Nanyang Technological University \ \ $^3$Stanford University\\
  \textcolor{magenta}{\url{https://lisiyao21.github.io/projects/Half-Physics/}} \\
}

\begin{document}

\maketitle

\begin{figure}[h]
    \centering
    \includegraphics[width=0.94\linewidth, trim=0pt 0pt 0pt 100pt, clip]{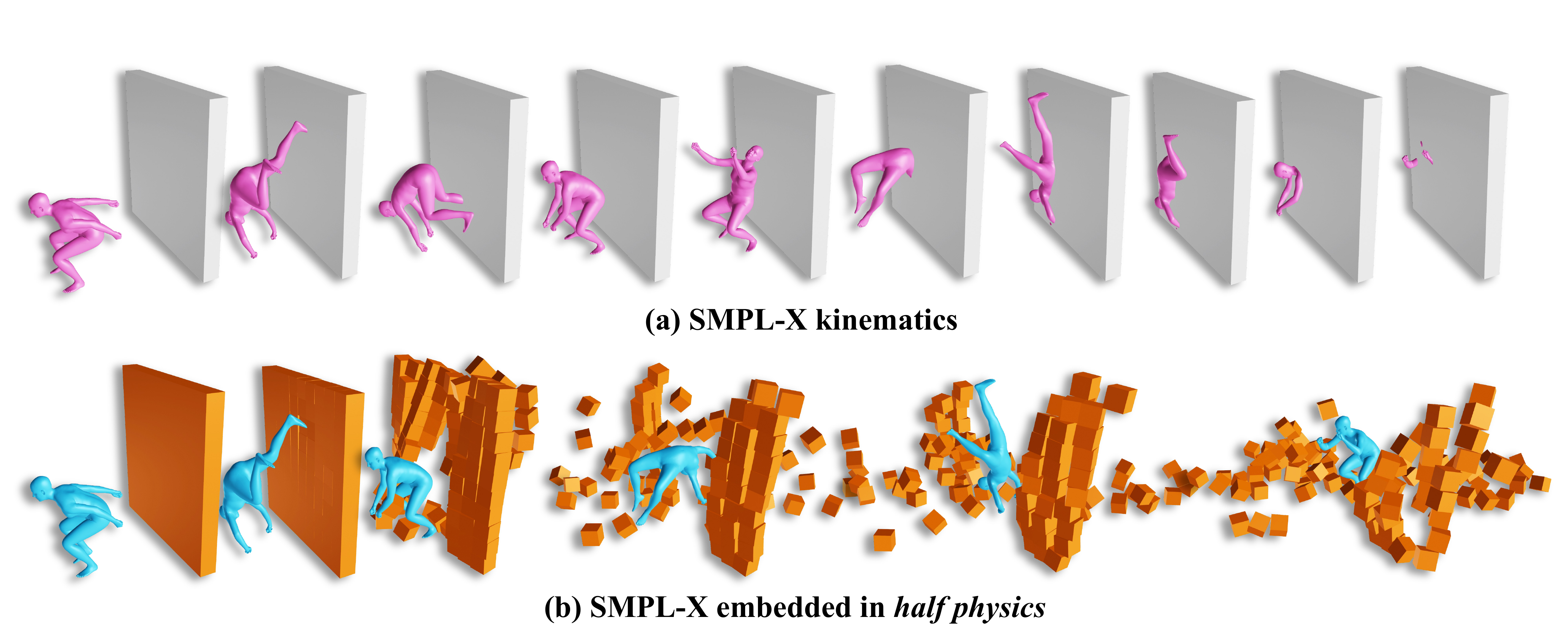}  
    \vspace{-10pt}
    \caption{\small\textbf{{Kinematic and \textit{half-physics} interactions of the SMPL-X 3D human model with the environment.} (a)} Standard kinematic SMPL-X meshes (\textcolor{black}{pink}) do not account for physical interactions, resulting in implausible penetration through obstacles. \textbf{(b)} Our proposed \textit{half-physics} framework (blue) allows the human model to produce physical interactions while keeping the original kinematic control.}
    \label{fig:teaser}
\end{figure}


\begin{abstract}
\vspace{-10pt}
While current general-purpose 3D human models (\eg, SMPL-X) efficiently represent accurate human shape and pose, they lacks the ability to physically interact with the environment due to the kinematic nature. 
As a result, kinematic-based interaction models often suffer from issues such as interpenetration and unrealistic object dynamics.
%
To address this limitation, 
we introduce a novel approach that embeds SMPL-X into a tangible entity capable of dynamic physical interactions with its surroundings. 
Specifically, we propose a ``\textbf{\textit{half-physics}}'' mechanism that transforms 3D kinematic motion into a physics simulation.
Our approach maintains kinematic control over inherent SMPL-X poses while ensuring physically plausible interactions with scenes and objects, effectively eliminating penetration and unrealistic object dynamics. 
%
%
Unlike reinforcement learning-based methods, which demand extensive and complex training, our \textit{half-physic}s method is learning-free and generalizes to any body shape and motion; meanwhile, it operates in real time.
Moreover, it preserves the fidelity of the original kinematic motion while seamlessly integrating physical interactions.
%



\end{abstract}

\vspace{-13pt}
\section{Introduction}
\vspace{-8pt}
\label{sec:intro}


Enabling 3D human models to physically interact with their environment is crucial for various applications, such as virtual reality (VR) and embodied AI.
%
Due to the intrinsic nature of existing human parametric models (\textit{i.e.},  SMPL \cite{SMPL:2015}, SMPL-H \cite{mano} and SMPL-X  \cite{smplx}) as purely kinematic mesh representations, they lack physical properties, making it difficult for them to interact realistically with the environment. 
%
%
As a result, interaction models often suffer from mesh penetration and physically implausible interactions (see Figure \ref{fig:teaser} (a)). 
Although  learning-based methods can incorporate loss functions \cite{hassan2019resolving,liu2024revisit} or rewards \cite{Zhao:ICCV:2023} 
to suppress these artifacts, they often struggle to generalize.

%
To integrate a  3D human model into a physics simulator, many existing methods employ reinforcement learning (RL) for imitation learning. In this setup, a virtual humanoid agent with joint motors is created inside the simulator. At each simulation timestep $\Delta t$, an RL policy predicts the torques (forces) for each joint so that the agent’s pose is driven towards the reference kinematic trajectory over time.
However, this approach has several drawbacks. 
\textbf{(1)} Computationally expensive. 
Imitation learning \cite{peng2018deepmimic,peng2021amp} usually requires extensive simulation time to model even a single action. 
\textbf{(2)} Lack of generalization. 
Even though recent methods \cite{luo2022embodied,luo2023perpetual,yuan2023learning,luo2024smplolympics} have improved single-action networks by making them more versatile, they still require separate dedicated training for each motion category (\eg, walking). 
Moreover, for a specific body shape, additional training is necessary to accurately track the reference motion \cite{luo2023perpetual}.
%
\textbf{(3)} Limited fidelity to kinematic data.
These methods control motion through per-frame impulses (forces applied over $\Delta t$), yet the agent already possesses momentum (velocity). 
This makes precise control difficult and often results in deviations from the original kinematic data (see Section \ref{sec:cmp2pt} and the supplementary video).

\begin{figure*}
   \centering
    \includegraphics[width=0.97\linewidth]{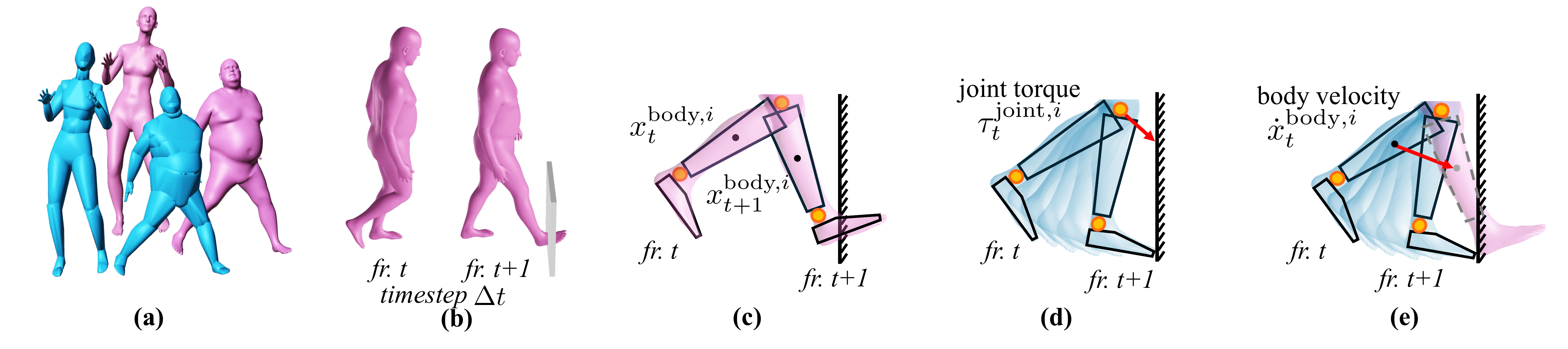}
    \vspace{-8pt}
    \caption{\small{
    \textbf{Illustration of the proposed \textit{half physics} frameword.}
    \textit{Half physics} bridges kinematics and physics by assigning equivalent velocities, enabling physically plausible transitions from purely kinematic inputs.
    \textbf{(a)} Shape-aware articulated variant (blue) of original human model (pink). 
    \textbf{(b)} {Two frames with kinematic poses}, with penetration at \textit{fr. t+1}. 
    \textbf{(c)} Kinematic poses $x^{\text{body},i}$ (for simplicity, we  depict only body positions here)  at discrete time $t$, physically intractable within $\Delta t$, and not able to block penetration.
    %
    \textbf{(d)} Robotics control uses torque indirectly affect velocity, are physically tractable during the timestep, and able to block penetration. 
    \textbf{(e)} Proposed \textit{half physics} explicitly infers body velocity towards kinematic target, physically tractable during the timestep, and able to block penetration. 
    }} 
        \vspace{-20pt}
    \label{fig:hp} 
\end{figure*}

In this paper, we propose a simple yet highly efficient and generalizable framework to extend kinematic human models to achieve physically plausible interactions with the environment while maintaining high fidelity to the original kinematic motion. 
%
%
%
%
%
%
Recall that it is challenging to integrate SMPL-X motions into physical simulations because they are defined by per-frame positions (including both 3D locations and joint rotations), yet these positions remain undefined within the continuous time interval  $\Delta t$.
To bridge this gap, rather than using raw position data, we compute the linear velocity of each body part from consecutive frames and explicitly apply it to the body state.
%
%
This simple yet effective approach enables \textit{seamless kinematic-to-physics transfer}: 
if no collision occurs with the environment, the human agent follows the predefined linear velocity to reach the target position in the next frame; 
if a collision happens, the response is governed by physical laws, ensuring realistic interactions.

The proposed approach is advantageous in that it keeps human motion kinematically driven by SMPL-X poses while allowing the surrounding scene and objects to be governed by physics, resulting in physically plausible interactions. We refer such a hybrid control mechanism as ``\textbf{\textit{half physics}}''. The proposed \textit{half-physics} framework requires no training and generalizes effectively to any body shape and motion. Moreover, by directly enforcing velocity instead of torque, it preserves kinematic fidelity while seamlessly integrating physical interactions (see Figure \ref{fig:teaser} and  the video).

%
To evaluate the usability of the proposed \textit{half-physics} approach, we conduct experiments on human-scene interaction (HSI) and human-object interaction (HOI) separately.
For HSI, we verify that 
\textit{half physics} eliminates penetration between the generated SMPL-X human body and the static environment. 
Notably, rather than preventing penetration through motion planning; \textit{half physics} allows the agent to perform the same motion as in the original kinematic sequence, with penetration naturally handled by the physics simulation (Section \ref{sec:hsi}).
For HOI, we demonstrate that within the half-physics framework, SMPL-X can interact with objects in various ways, including single-handed grasping, two-handed lifting, dropping, and kicking.
Additionally, we show that \textit{half physics} can be used to augment existing interaction datasets by applying different physical factors, enriching interaction outcomes with more diverse physical labels (Section \ref{sec:hoi}).
Across all experiments, \textit{half physics} takes SMPL-X as input, not only eliminating penetration but also enabling seamless physical interactions, extending the capabilities of kinematic-centric SMPL-X.


\vspace{-6pt}
\section{Related Work}
\label{sec:related}
\vspace{-5pt}




\noindent{\textbf{Human Parametric Models.}}
The SMPL family \cite{SMPL:2015,mano,smplx} provides a series of efficient human body representations to precisely control human shape ($\beta$) and pose ($\theta$) using only a small number of parameters. 
These models have been widely used in human-centric research, including human pose estimation and motion capture \cite{bogo2016keep,kanazawa2018end,kocabas2020vibe,saini2022airpose,feng2024chatpose}, motion generation \cite{li2021ai,siyao2022bailando,siyao2023bailando++,siyao2024duolando}, human reconstruction \cite{xiu2022icon,xiu2023econ},
%
and 3D avatar generation \cite{hong2022avatarclip}, 
enabling the creation of detailed and expressive 3D assets for various content generation tasks. 
%
%
However, SMPL-X remains a purely kinematic model, lacking physical properties necessary for realistic 
interaction with the environment.

\noindent{\textbf{Motion Imitation.}}
Motion imitation aims to enable humanoid robots or character agents, driven by joint motor torques, to replicate human kinematic motions while maintaining physical plausibility in simulation. 
Peng \textit{et al.} \cite{peng2018deepmimic,peng2021amp} and the following works \cite{luo2020carl,ren2022diffmimic} introduce early approaches requiring case-by-case training. 
In recent years, efforts have been made to improve generalization, allowing models to learn a broader range of motions \cite{luo2022embodied,wagener2022mocapact} and to be applied to various tasks, such as sports \cite{luo2024smplolympics,yuan2023learning,wang2025skillmimic,luo2025humanoidlympics}.
More recently, Luo \textit{et al.} \cite{luo2023perpetual,luouniversal} propose a more generalized physical tracker. 
However, aside from the complexity of its training process, it still exhibits centimeter-level discrepancies from the target kinematics, making it unsuitable for kinematically driven interactions.

\noindent{\textbf{Human Scene Interaction (HSI).}}
%
%
Most recent HSI work \cite{araujo2023circle,hassan2021stochastic, huang2023diffusion, mir2023generating, pan2021synthesizing, wang2021synthesizing, wang2021scene, wang2022towards, xu2023interdiff,jiang2024scaling} leverages the SMPL-X body representation and learns human behaviors in environments from large-scale datasets. 
%
%
However, due to the kinematic nature of SMPL-X, it cannot inherently prevent human-scene penetrations, leading to unrealistic interactions.
Some methods mitigate this issue by introducing a penetration loss \cite{hassan2019resolving} or precomputing obstacle-free paths \cite{jiang2024autonomous}, but these approaches do not generalize well.
Several works \cite{xiao2024unified,tevet2025closd} learn interaction patterns within physics simulators to fundamentally avoid penetration issues. 
However, these methods typically rely on imitation learning of a joint-motor-driven humanoid, leading to the above-mentioned drawbacks.
In contrast, our proposed \textit{half-physics} approach enables kinematic predictions to maintain motion fidelity while blocking penetrations, ensuring realistic human-scene interactions.

\noindent{\textbf{Human Object Interaction (HOI).}}
Kinematic-based HOI models \cite{li2023object,li2024controllable,wu2024human} predict both human and object motion patterns from MoCap interaction datasets, often leading to implausible interactions. 
On the other hand, physics-based approaches \cite{tevet2025closd,luo2024omnigrasp,wang2023physhoi,xu2025intermimic} rely on motion imitation. 
Again, this makes them ill-suited for HOI scenarios that require kinematic control, such as VR applications.
In contrast, the proposed \textit{half physics} enables physically plausible interactions while maintaining SMPL-X kinematic control.


\vspace{-5pt}
\section{Method}
\vspace{-5pt}
\label{sec:method}


Our goal is to enable the SMPL-X human models to interact physically with the environment.
To achieve this, we aim to efficiently and seamlessly integrate SMPL-X into a physics engine.
However, SMPL-X is inherently a kinematic model, meaning that human motion is represented without physical dynamics.
Typically, a motion sequence $\bm \theta$ of SMPL-X is defined as a 3D global translation (position) $\bm x^{\text{global}} \in \mathbb R^{T\times3}$, a 3D global orientation $\bm q^{\text{global}} \in \mathbb R^{T\times3}$, and joint rotations $\bm q^{\text{joint}} \in \mathbb R^{T\times J \times3}$, where $T$ is the frame number and  $J=54$ is the joint number.
In this section, we elaborate on the gap between SMPL-X and physics, and propose a simple yet effective velocity-based dynamics implementation to integrate SMPL-X into a physics engine.

In all subsequent notations, $x$ represents a 3D position
, while $q$ is 3D rotation (represented in quaternion);
superscripts (\eg, $^{\text{joint},i}$) denote the associated object  and index of a physical quantity (\eg, the $i$-th joint), subscripts $_t$ represent time point (frame), while dot $\dot{}$ and double dot $\ddot{}$ over a variable  indicates its first and second derivatives (velocity and acceleration), respectively.
%
%
%


\vspace{-3pt}
\subsection{Gap between Kinematic Models and Physics}
\vspace{-3pt}
The main challenge in integrating SMPL-X into physics engines lies in its pose being specified only at discrete time points (frames). 
As a result, the human state within the continuous in-between interval is indeterminate, making it impossible for the physics engine to process interactions during these periods.
An example is illustrated in Figure \ref{fig:hp}(b), where the human does not penetrate at frame $t$, but instantaneously does at frame $t+1$.
A closer examination is shown in Figure \ref{fig:hp}(c), where the body is at position $x^{\text{body},i}_t$ in the previous frame but has already moved to $x^{\text{body},i}_{t+1}$, penetrating the wall in the next frame, with no intermediate transition.
Since no state exists between these two frames, the simulation fails to register the collision and consequently cannot prevent the penetration.

\noindent{\textbf{Conventional Approaches.}}
To track character
motion in the continuous time domain, existing reinforcement learning (RL) approaches, which are called ``\textit{physical trackers}'', typically predict torques, \ie, the force to twist joint, at each frame (see Figure \ref{fig:hp}(d)). 
During simulation, these torques are converted into accelerations, which are then integrated to velocity updates for each body part. 
Over the entire interval $\Delta t$, velocity remains continuously tractable, allowing the simulator to respond to collisions physically. 
However, controlling a human agent like a robot is complex, as it requires accounting for the inertia of each link from past actions, often leading to instability, \eg, falling (see Section \ref{sec:cmp2pt}).
%
Moreover, simulating human-centric interactions does not necessitate building a humanoid robot in ``\textit{full physics}'' from scratch, 
as long as its dynamics are compatible with the physics engine.
%
Therefore, we propose a more direct approach that explicitly {\em manipulates velocity rather than torque}, providing a smoother transition from kinematics to the physical engine while better aligning with SMPL-X human motion.
%
\vspace{-3pt}
\subsection{Our Solution}
\vspace{-4pt}
\noindent{\textbf{SMPL-X Variant.}}
Similar to Luo~\etal~\cite{luo2022embodied}, we construct an articulated rigid-body model that approximately matches the shape of the SMPL-X model (see Figure \ref{fig:hp}(a), blue), which is called a \textit{SMPL-X variant}. 
Specifically, given an input shape parameter $\beta$, we use the segmentation labels to partition the vertices into distinct components and enclose them as rigid body parts. 
%
This variant consists of $55$ rigid body parts, each of which (indexed as $i$) has its own position $x^{\text{body,}i}$ and orientation $q^{\text{body,}i}$.
The pose of the variant is aligned to SMPL-X, determined by a global translation $x^{\text{global}}=x^{\text{body,}0}$, a global rotation $q^{\text{global}}=q^{\text{body,0}}$ and $54$ joint angles $\bm{q}^{\text{joint}}$. 
It is worth noting that, unlike motion imitation, which aims to drive an agent to perform similar actions, our method focuses on giving SMPL-X a physical presence while executing the same motion.

{
\begin{algorithm}[t]
\small
\begin{algorithmic}[1]
        \Statex \textbf{Input} static scene $s$, initial object state \{$x^{\text{obj}}_0$, $q^{\text{obj}}_0$\},  SMPL-X pose sequence $\bm \theta$, shape $\beta$, timestep $\Delta t$
        
        \Statex \textbf{Output} object motion \{$\hat{\bm x}^{\text{obj}}$, $\hat{\bm q}^{\text{obj}}$\},  corrected poses $\hat{\bm \theta}$
    \State  $\operatorname{\textcolor{cyan}{\texttt{sim}}}\gets \operatorname{Simulator}(s,\beta)$ {\small{\LightComment{Create physical simulator}}}
    \State $\hat{\bm q}^{\text{joint}}_0, \hat x^{\text{body},0}_0, \hat q^{\text{body},0}_0\gets\bm\theta_0$
    {\small{\LightComment{Initialize agent state}}}
    \State $\hat x^{\text{obj},0}_0 \gets x^{\text{obj},0}_0, \,\, \hat q^{\text{obj},0}_0 \gets q^{\text{obj},0}_0$
    {\small{\LightComment{Initialize object state}}}

        \For{frame $t \gets 1$ to $T-1$}
            \State ${\bm q}^{\text{joint}}_t, x^{\text{body},0}_t, q^{\text{body},0}_t\gets\bm\theta_t$ 

            \State $\dot{\bm q}^{\text{joint}}_t \gets \operatorname{angdiff}({\bm q}^{\text{joint}}_{t}, \hat{\bm q}_{t-1}^{\text{joint}})/\Delta t$ {\small{\LightComment{Joint velocities}}}
            
            \State $\dot{x}^{\text{body},0}_t \gets  (x^{\text{body,0}}_t - x^{\text{body,0}}_{t-1})/\Delta t $
            {\small{\LightComment{Global linear velocity}}}
            
            \State $\dot{q}^{\text{body},0}_t \gets \operatorname{angdiff}( q^{\text{body},0}_t, \hat q_{t-1}^{\text{body},0})/\Delta t$ 
            {\small{\LightComment{Global angular velocity}}}


            \State $\ddot{x}^{\text{obj}}_t,\ddot{q}^{\text{obj}}_t\gets\operatorname{gravity}$
            %
            {\small{\LightComment{Apply gravity on object}}}
            
            \State $\hat {\bm x}^{\text{body}}_t, \hat{\bm q}^{\text{body}}_t, \hat x_t^{\text{obj}}, \hat q_t^{\text{obj}} 
            \gets \operatorname{\texttt{\textcolor{cyan}{sim}}}({\Delta t})$
            {\small{\LightComment{Simulate}}}

            \State $\hat{\bm \theta}_t \gets \hat{\bm q}^{\text{joint}}_t,\hat x^{\text{body},0}_t, \hat q^{\text{body},0}_t$
        \EndFor
        \State \textbf{return} $\hat{\bm \theta}, \hat{\bm x}^{\text{obj}}, \hat{\bm q}^{\text{obj}}$ 
\end{algorithmic}

\caption{\small{Half-Physics Simulation $\mathcal H$}}
\label{alg:hp}
\end{algorithm}
}

\noindent{\textbf{Half Physics.}}
With the variant aligned to the original SMPL-X mesh, we transform SMPL-X kinematic poses into dynamics by \hp.
%
The term ``\hp'' refers to a setting in which the human is driven by kinematic poses, while objects adhere to physical laws, distinguishing it from RL-based approaches that rely on joint motor actuation to power movement.
%
%
%
%
Instead of applying torques, which indirectly control velocity, we explicitly enforce a velocity for each body part at every time point $t$. 
As shown in Figure \ref{fig:hp}(e), the velocity is intuitively approximated as the finite difference between the kinematic target position and the current frame:
$\dot x_t^{\text{body},i} = \frac{x_{t+1}^{\text{body},i} - x_{t}^{\text{body},i}}{\Delta t}$, which results in the body part reaching the target position $x_{t+1}^{\text{body},i}$ over the duration of $\Delta t$.
%
%
%
%
By enforcing velocities, we achieve continuous tractability, making correct  collision responses and  blocking penetration.

The general algorithm for \textit{half physics} is provided in Algorithm \ref{alg:hp}. 
%
%
In general, given a specific scene $s$ and an initial object state, \textit{half physics} simulates the interaction between the human and the surroundings, ultimately producing a physically corrected human SMPL-X pose sequence and the object trajectory:
\begin{equation}
    \hat{\bm \theta}, \hat x^{\text{obj}}_{1...T},  \hat q^{\text{obj}}_{1...T} = \mathcal{H}^s(\bm \theta;x^{\text{obj}}_{0}, q^{\text{obj}}_{0}).
\end{equation}
At each frame $t$, \textit{half physics} treats $\bm \theta_t=\{\bm q^{\text{joint}}_t, x^{\text{body},0}_t, q^{\text{body},0}_t\}$ as the motion target and progressively simulates to obtain $\hat{\bm \theta}_t$, $\hat x^{\text{obj}}_t$ and $\hat q^{\text{obj}}_t$. 
%
%
Joint velocities $\dot{\bm q}^{\text{joint}}_t$ and the global angular velocity $\dot q_t^{\text{body, 0}}$ are computed via spherical linear interpolation \cite{slerp} (SLERP, noted as ``angdiff'')  between their targets ($\bm q^{\text{joint}}_t$ and $q_t^{\text{body, 0}}$) and their current states ($\hat{\bm q}^{\text{joint}}_{t-1}$ and $\hat q_{t-1}^{\text{body, 0}}$), respectively.
In contrast, the global linear velocity  $\dot x_t^{\text{body, 0}}$ is computed as the difference between the target global positions at frames $t$ and $t-1$.
%
%
Then, physics simulation is executed to step an increment of $\Delta t$.
As part of this process, joint angular velocities $\dot{\bm q}^{\text{joint}}$ 
are used to compute body part velocities $\dot{\bm x}^{\text{body}}_t$ and $\dot{\bm q}^{\text{body}}_t$ via forward kinematics, while inverse kinematics is applied to infer the updated joint states from the new body states after the simulation step.
If no collision occurs, the \textit{half-physics} SMPL-X variant can maintain the same pose sequence as the original SMPL-X, \ie, $\hat{\bm \theta} =\mathcal H(\bm \theta)$, thereby achieving a faithful transfer to physics simulation.
Unlike RL-based physical trackers, $\hat{\bm \theta}$ from \textit{half-physics} preserves the original motion with no loss in fidelity (see Section \ref{sec:cmp2pt}). 
Note in HOI tasks, \textit{half physics} uses only the first-frame object pose for \textit{initialization}; all subsequent object trajectories are generated purely through simulation, without any predefined motion. 
This is fundamentally different from preprocessing-based methods \cite{wang2023physhoi,xu2025intermimic}, which input both human and object poses independently at each frame and use the simulator only to resolve penetration. 
In contrast, our approach produces physically consistent object responses that emerge naturally from interaction, rather than being imposed frame-by-frame.
Implementation details of \texttt{\textcolor{cyan}{sim}} are provided in Appendix~\ref{sec:impl_detail}.

\section{Experiments}
\label{sec:experiment}

   
 


The proposed \textit{half physics} can efficiently transform kinematic SMPL-X into a physics simulator, enabling physical interactions. 
In this section, we conduct three experiments to validate and explore the effectiveness of \textit{half physics}: \textbf{(1)} comparing with state-of-the-art RL-based physical tracker PHC+ \cite{luouniversal}, \textbf{(2)} eliminating SMPL-X penetration in human-scene interaction (HSI), and \textbf{(3)} exploring the interactive behaviors of \textit{half physics} in human-object interactions (HOI).

\begin{figure}[t]
    \begin{minipage}{0.49\linewidth}
        \includegraphics[width=0.88\linewidth]{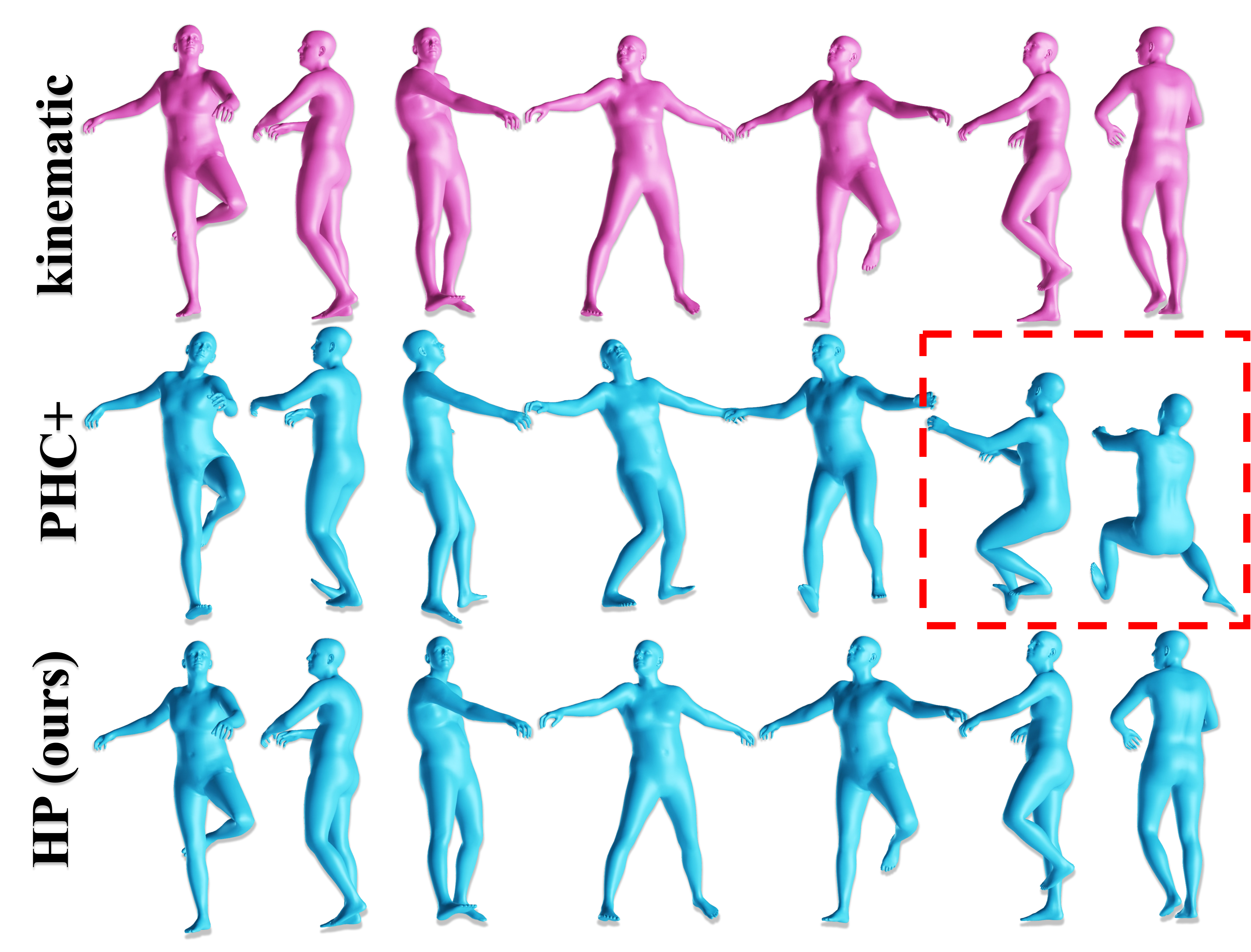}
    \vspace{-4pt}
    \caption{\small\textbf{Qualitative comparison to PHC+ \cite{luouniversal}.} The tracked subject falls over PHC+  (red box), while ours keeps high fidelity to kinematic reference.}
    \label{fig:comp2phcp}
    \end{minipage}
 \begin{minipage}{0.479\linewidth}


\centering
  \centering
    \captionof{table}{
\small{
\textbf{Comparison to state-of-the-art physical tracker PHC+ \cite{luo2023perpetual} on AIST++ test set \cite{li2021ai}}.
$^*$Computed on successful samples.
}}
    \scriptsize{
  \centering
  {
  \begin{tabular}{ l   c  c c c  }
    \toprule
\scriptsize
   Method & Succ. ($\%$) $\uparrow$  & MPJPE-\textit{g} (\textit{mm})$\downarrow$ & MPJPE $\downarrow$\\
   
\midrule
 PHC+ \cite{luo2023perpetual} & $\,\,\,\,92.5$  & $49.192^*$ &  $34.467^*$  \\
HP (ours) & $\bf 100.0$ & $\,\bf0.003$  &  $\,\bf 0.003$\\
    \bottomrule
  \end{tabular}
  }
  }
  \vspace{2em}  


    \centering
    \captionof{table}{\small{\textbf{Quantitative evaluations on Trumans}. Depths are measured in \textit{mm}.}}
    \small{
  \centering\scriptsize{
  \begin{tabular}{ l  c c c}
    \toprule

   Method &  Pene ($\%$) $\downarrow$ &  Depth$_{mean}$ $\downarrow$ &  Depth$_{max}$ $\downarrow$\\
\midrule
 
 Trumans \cite{jiang2024scaling} &  $7.91$ & $82.10$ & $172.19$ \\
 
Trumans + HP & $\bf 0.00$& $\,\,\,\bf0.00$ & $\,\,\,\,\,\,\,\bf0.00$ \\
    \bottomrule
  \end{tabular}
  }
  
  
  \label{table:trumans-hp}
  }

\label{tab:cmp2phcp}
    \end{minipage}

    \vspace{-5pt}
\end{figure}
\vspace{-3pt}
\subsection{Comparison to  Physical Tracker PHC+ \cite{luouniversal}}
\vspace{-3pt}
\label{sec:cmp2pt}


A physical tracker aims to make a humanoid robot imitate a target kinematic motion in physics simulation.
Recently, PHC \cite{luo2023perpetual} and PHC+ \cite{luo2023perpetual} have made significant progress in improving tracking generalization.
To test the effectiveness of our \textit{half-physics} approach, we compare its tracking performance with PHC+. 
Specifically, we apply both \textit{half physics} and PHC+ to track $40$ 3D dance sequences from the AIST++ test set. 
We calculate the success rates, defined as the proportion of sequences without interruptions by falling, along with the global mean per-joint point error (MPJPE-\textit{g}), which is the average distance between the tracked joint positions and the target positions, and local MPJPE, which considers only the error relative to the root joint. 
%
The AIST++  dataset  contains  many intricate, fluid, and dynamic human dance motions, and hence is difficult to imitate by robotics. 
Moreover, AIST++ is reconstructed from multi-view data, not from dedicated MoCap data, making it more challenging. 
We use the PHC+ ``\texttt{rotation+keypoint}'' version, which reports a $100\%$ success rate on AMASS.
For a fairer comparison, we finetune PHC+ for $5$ epochs  on the AIST++ training set.

The results are shown in Table \ref{tab:cmp2phcp}. 
PHC+ achieves a success rate of $92.5\%$, with 3 sequences resulting in interruptions by falling (see visualization in Figure \ref{fig:comp2phcp}). 
In contrast, \textit{half physics} (HP) completes $100\%$ of the sequences.
Regarding MPJPE-\textit{g}, PHC+ has an average tracking error distance of nearly $5$ \textit{cm} (approximately the width of three fingers).
Note that these results are based only on successful samples; including failure cases would further widen the gap.
In contrast, HP achieves a global error of merely $3$ \textit{$\mu$m} (roughly the length of an \textit{E. coli} bacterium), which is virtually negligible.
For local MPJPE, which better reflects pose tracking, PHC+ has an error greater than $3$ \textit{cm}, about $70\%$ of the MPJPE-\textit{g} error, indicating that maintaining pose consistency in dynamic dance motions is the primary challenge for its imitation; 
while HP achieves a local MPJPE of $3$ \textit{$\mu$m},  identical to its global MPJPE-g.
The remaining  errors in HP primarily stem from rounding errors during velocity computation.
Compared to state-of-the-art physical trackers, HP  provides a significant advantage, reducing the error from the centimeter scale to a negligible one, and hence making it highly promising for interactive applications such as virtual reality.

For training, PHC+ needs $7$ days and requires complex RL strategies, along with reward engineering tailored to different motions. 
In contrast, \textit{half physics} is {\em training free}.
Regarding inference time, \textit{half physics} runs 
$952.94$
fps on average in this experiment on a desktop  with Intel Xeon CPU E3-1280 V2 @ 3.60GHz and Nvidia Quadro RTX 5000 GPU (16G RAM), far over a real-time speed.

\begin{figure*}[t]

    \setlength{\tabcolsep}{1.5pt}
    \centering
    \includegraphics[width=0.97\linewidth]{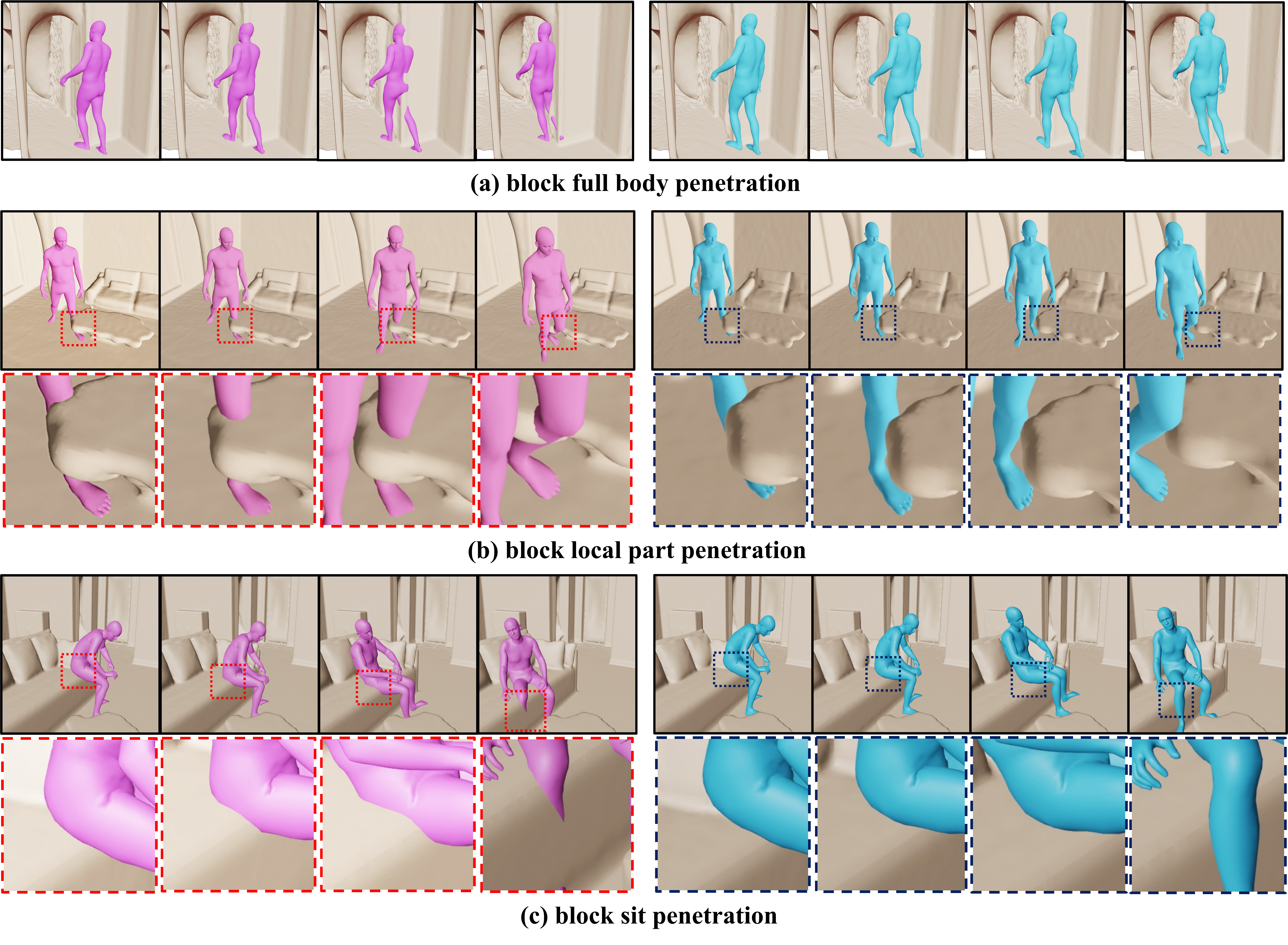}

    \vspace{-8pt}
    \caption{\small{\textbf{
    Qualitative results of Trumans \cite{jiang2024scaling} (pink) and Trumans+HP (ours) (blue).
    }
    The proposed \textit{half physics} can effectively eliminate different kinds of penetration in human scene interactions while maintain high fidelity to original kinematic motion. 
    } }
        \vspace{-5pt}
    \label{fig:trumans}

\end{figure*}

\vspace{-5pt}
\subsection{Human Scene Interaction}
\vspace{-5pt}
\label{sec:hsi}

Penetration in human-scene interaction remains a critical challenge, affecting both visual realism and accuracy in embodied AI. 
Common solutions include post hoc optimization with a penetration loss \cite{hassan2019resolving} or using motion planning to precomputed the navigation route \cite{jiang2024autonomous}. 
However, these methods often fail to generalize and may constrain humanoid movement (\eg, nav-meshes typically prevent actions like sitting on a bed or sofa).
In contrast, by embedding SMPL-X into \textit{half physics}, the human agent can still perform the full range of motions while physical penetration is blocked.



We evaluate our method on the Trumans \cite{jiang2024scaling} human-scene interaction dataset, demonstrating its effectiveness in resolving penetration issues.
Specifically, we follow Jiang \etal \cite{jiang2024scaling} to generate SMPL-X motion sequences $\bm{\theta}$ in $20$ diverse test scenes $\{s\}$ and use them as inputs to the Algorithm \ref{alg:hp} to obtain corrected motions:
\begin{equation} 
\small{\hat{\bm{\theta}} = \mathcal{H}^s (\bm{\theta})}. 
\end{equation}
In this experiment, since these scenes do not contain movable objects, our evaluation focuses solely on penetration resolution.
As for quantitative metrics, we compute the average penetration rate ($\%$, noted as ``Pene'') of SMPL-X vertices, average penetration depths (\textit{mm}) per frame (noted as ``Depth$_{mean}$'') and per sequence (noted as ``Depth$_{max}$'') between human and scene meshes.

As shown in Table \ref{table:trumans-hp}, the motions processed by the \textit{half-physics} simulation can  completely eliminate penetration in numerical terms, eliminating the penetration rate from $7.914\%$ to 
\textit{zero}.
%
In terms of penetration depth, the original generated motion $\bm \theta$ consists of purely kinematic motions, exhibiting an average penetration depth of approximately $8$ \textit{cm} per frame, with the maximum penetration within a sequence reaching $17$ \textit{cm}—roughly the length of a human hand. 
In contrast, after applying the same kinematic data to the \textit{half-physics} simulation, the penetrations are effectively eliminated at their root.


To provide deeper insight into the results, we visualize three examples in Figure \ref{fig:trumans}.
In the first example (Figure \ref{fig:trumans}{(a)}), we demonstrate a case of full-body penetration. 
The human motion generated by Trumans (left, colored in pink) shows the right half body passing through a wall while walking through a doorway; 
in contrast, after undergoing \textit{half-physics} simulation (right, colored in cyan), the human is physically blocked from proceeding further.
Figure \ref{fig:trumans}{(b)} illustrates the penetration of a local body part, where the original motion generated by Trumans causes the leg to pass through a coffee table. After applying \textit{half-physics}, the leg is properly obstructed, effectively eliminating penetration.
Furthermore, we show a common challenge in human-scene interaction, where the human sinks into a sofa or bed when performing a sitting action (see Figure \ref{fig:trumans}{(c)}). The \textit{half-physics} simulation successfully removes such sinking artifacts.
Noticeably, beyond eliminating penetration, the motion corrected by \textit{half-physics} remains faithful to the original intent of the motion.
This outcome is not due to an \textit{active avoidance} of objects but instead a natural enforcement of physical constraints that block penetration. 
This helps generalize human motion to new scenes by avoiding penetration, illustrating that our \textit{half physics} approach provides a foundation for advancements in embodied AI applications.
Additional results can refer to the \href{https://lisiyao21.github.io/projects/Half-Physics/}{\textcolor{magenta}{\texttt{video}}}.

\begin{figure*}[t]

    \setlength{\tabcolsep}{1.5pt}
    \centering
    \includegraphics[width=0.95\linewidth]{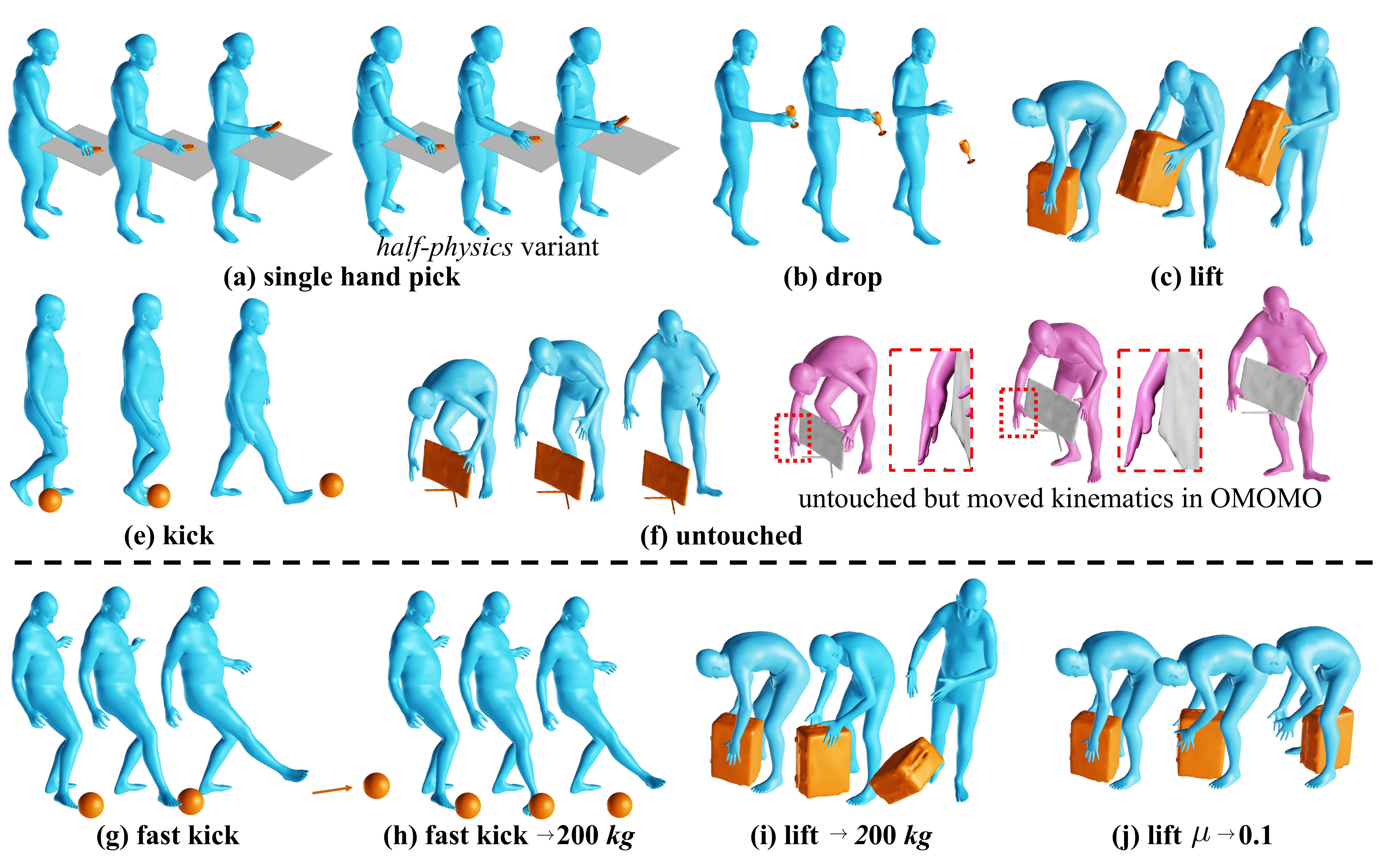}

    \vspace{-8pt}
    \caption{\small{\textbf{
    \textit{Half physics} can enable kinematic human models to perform diverse interactions with objects. %
    } \textbf{(a-e)} Human motions are \emph{kinematic} while objects (orange) are driven by physics. 
    \textbf{(f)} Without contact, pure kinematic models generate unrealistic  object (gray) moving, while \textit{half physics} avoids this.
    \textbf{(g-j)} The amplitude of kinematic motion and physical factors of objects further augment the interaction outcomes.}} 
        \vspace{-15pt}
    \label{fig:hoi}

\end{figure*}

\vspace{-5pt}
\subsection{Human Object Interaction}
\label{sec:hoi}
\vspace{-5pt}

Producing plausible object responses to kinematic human motion is challenging in an interactive system.
On one hand, kinematic data-driven (or, SMPL-X-based) methods typically rely on deep learning networks to \textit{predict} object motions \cite{li2023object,li2024controllable,wu2024human}, which often leads to unrealistic behaviors.
For example, objects may start moving before any actual contact with humans. 
Moreover, such predictions are prone to overfitting -- 
if the training data lacks specific interactions, \eg, object dropping, the model may fail to generate such motions, resulting in unrealistic human-object interactions.
On the other hand, purely physics-based approaches, such as RL-based imitation, face the previously discussed challenges, including complex training, limited generalization, and limited fidelity.
This may impose limitations on applications that strongly rely on kinematic input, such as VR.
%
With the help of \textit{half physics}, there is no need to predict object motion through additional learning methods. 
By using the object's initial state  $\{x^{\text{obj}}_0, q^{\text{obj}}_0\}$ and the SMPL-X pose sequence $\bm \theta$, \textit{half physics} can directly produce physically plausible interaction results:
\begin{equation}
\small{
    \hat{x}^{\text{obj}}_{1...T}, \hat{q}^{\text{obj}}_{1...T} = \mathcal{H}(\bm\theta; x_0^{\text{obj}}, q_0^{\text{obj}}).
    }
\end{equation}
In Figure \ref{fig:hoi}, we present a series of human-object interactions realized through the proposed \textit{half physics}. 
We input existing interaction data (\eg, from GRAB \cite{GRAB:2020} and OMOMO \cite{li2023object}) into the half-physics algorithm. 
%
%
%
Figure \ref{fig:hoi}(a) shows that, with the proposed \textit{half physics}, the agent can physically grasp an object with a single hand through direct control of SMPL-X. 
It is worth noting that in Figure \ref{fig:hoi}, all orange-colored objects are \textit{physics-driven}, meaning their motions are not dictated by kinematic trajectories but rather emerge from physical interactions with SMPL-X variant (see Figure \ref{fig:hoi}(a), right part). 
%
%
Next, Figure \ref{fig:hoi}(b) demonstrates the process of releasing a wine glass, causing it to fall. 
In the original data, the human intends to place the wine glass on a table. 
However, since we withdraw the table in our simulation, the glass naturally falls due to gravity. 
This example highlights how \textit{half-physics} can generate realistic but uncommon interaction dynamics beyond those explicitly present in kinematic MoCap dataset.
Figure \ref{fig:hoi}(c) shows a scenario where the human lifts a suitcase by frictional forces between both hands. 
This demonstrates that \textit{half-physics} can support full-body manipulative interactions, enabling physically plausible transportation tasks.
Furthermore, we explore whether the \textit{half physics} can achieve plausible SMPL-X kicking behaviors.
In our simulation, we place a ball at an appropriate position to interact with a kicking motion derived from the SMPL-H data in AMASS \cite{mahmood2019amass}. 
As shown in Figure \ref{fig:hoi}(e), the ball is naturally kicked forward as the human moves, demonstrating the potential of \textit{half-physics} for research and applications related to sports motion.
Going a step further, we investigate whether different kicking styles result in correspondingly distinct ball movements. 
Figure \ref{fig:hoi}(g) shows an example of a large-magnitude, fast kick, where the ball is forcefully propelled forward and lifted into the air, confirming that \textit{half-physics} enables motion-magnitude-aware physical interactions. 
This enhances the realism of human-object interactions in dynamic motion scenarios.
Finally, we examine cases where no contact occurs. 
On the right side of Figure \ref{fig:hoi}(f), we present an example from the OMOMO \cite{li2023object} kinematic interaction dataset, where the human does not touch the monitor, yet the object still moves. 
Such interactions are inherently unrealistic. 
In contrast, under \textit{half physics}, if no contact occurs, the monitor remains unaffected and is not lifted (left side, Figure \ref{fig:hoi}(f)).
This example further demonstrates the potential of \textit{half-physics} as a platform for human-object interaction synthesis, ensuring physically plausibility.

\noindent\textbf{Augmenting HOI Data with Physical Factors.} 
In real-world interactions, not only the shape of objects matters but also various physical factors such as mass and friction. 
Understanding human interactions under different physical conditions is highly beneficial for event prediction and action decision-making in embodied AI and robotics.
However, due to real-world constraints—such as the difficulty of constructing identical objects with different physical properties—such data are challenging to obtain through motion capture (MoCap).
Here, we present another application of \textit{half physics}: augmenting human-object interaction (HOI) data with physical factors. 
Figures \ref{fig:hoi}(h–j) illustrate interactions under different object masses and friction coefficients ($\mu$).
In Figure \ref{fig:hoi}(h), we increase the ball's mass from $500$ \textit{g} (approximated weight of a football) to $200$ \textit{kg}. 
Under the same kicking motion, the ball remains nearly static. 
Notably, due to the excessive resistance, the simulated SMPL-X posture in \textit{half-physics} adjusts dynamically: the right foot bends and alters its trajectory and merely glances off the ball instead of going through it. 
This occurs because the human body's momentum is physically corrected upon collision, leading to more realistic interactions.
Similarly, when increasing the suitcase’s weight to $200$ \textit{kg}, the human is unable to lift it. 
Additionally, the fingers bend under strain, reflecting a physically plausible reaction (see Figure \ref{fig:hoi}(i)). 
Since lifting requires sufficient friction, when the suitcase’s friction coefficient is reduced to $0.1$, the suitcase slips away under hand pressure (see Figure \ref{fig:hoi}(j)).
The interaction results under different physical factors—whether aligned or misaligned with human motion intention—can serve as prior knowledge to enhance action generation in embodied AI and robotics.
This further demonstrates the potential usage of the proposed \textit{half-physics} approach in synthesizing physically-aware human-object interactions.


\begin{figure}[t]
\begin{minipage}{0.43\textwidth}
    \centering
    \includegraphics[width=0.99\linewidth]{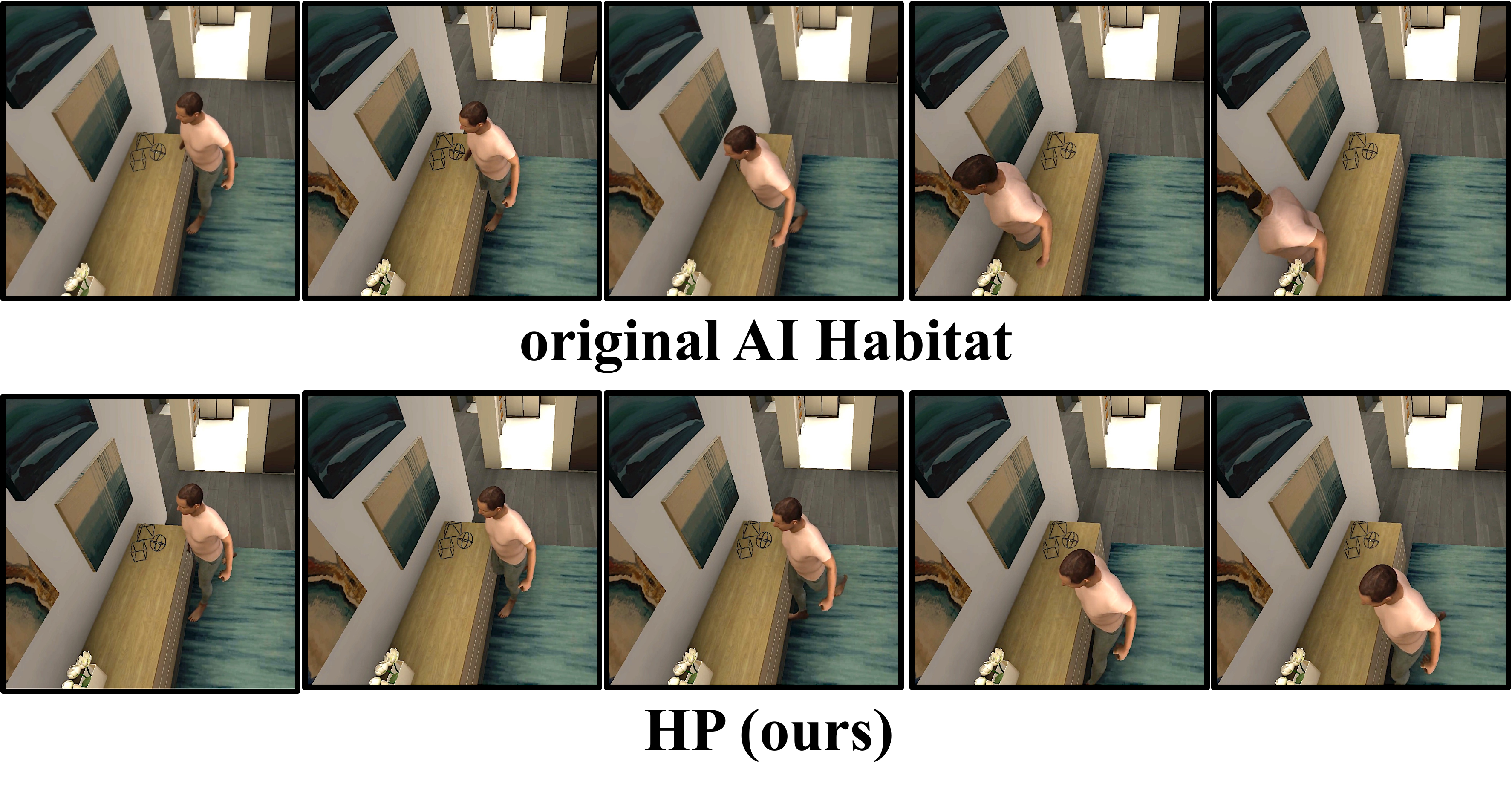}
    \caption{\small\textbf{Comparison on original AI Habitat \cite{puig2023habitat} and our implemented \textit{half physics} (HP).}}
    \label{fig:ablation1}
    \vspace{-15pt}
\end{minipage}
\begin{minipage}{0.53\textwidth}

    \centering
    \small\captionof{table}{
    {
    \textbf{Ablation study on vel. override in HP}
    }}
      \label{table:ablation2}
      \vspace{-5pt}
    \centering
    \small{
      \centering\scriptsize{
      \begin{tabular}{ l    c c  }
        \toprule
       Method   & MPJPE-\textit{g} (\textit{mm})$\downarrow$ & MPJPE $\downarrow$\\
    \midrule
    HP vel.$\rightarrow$torque & $83.109$  &  $82.273$\\
     HP  & \ \ $\bf0.003$  &  \ \ $\bf 0.003$\\
        \bottomrule
      \end{tabular}
      }
      }

    \centering
    \includegraphics[width=0.73\linewidth]{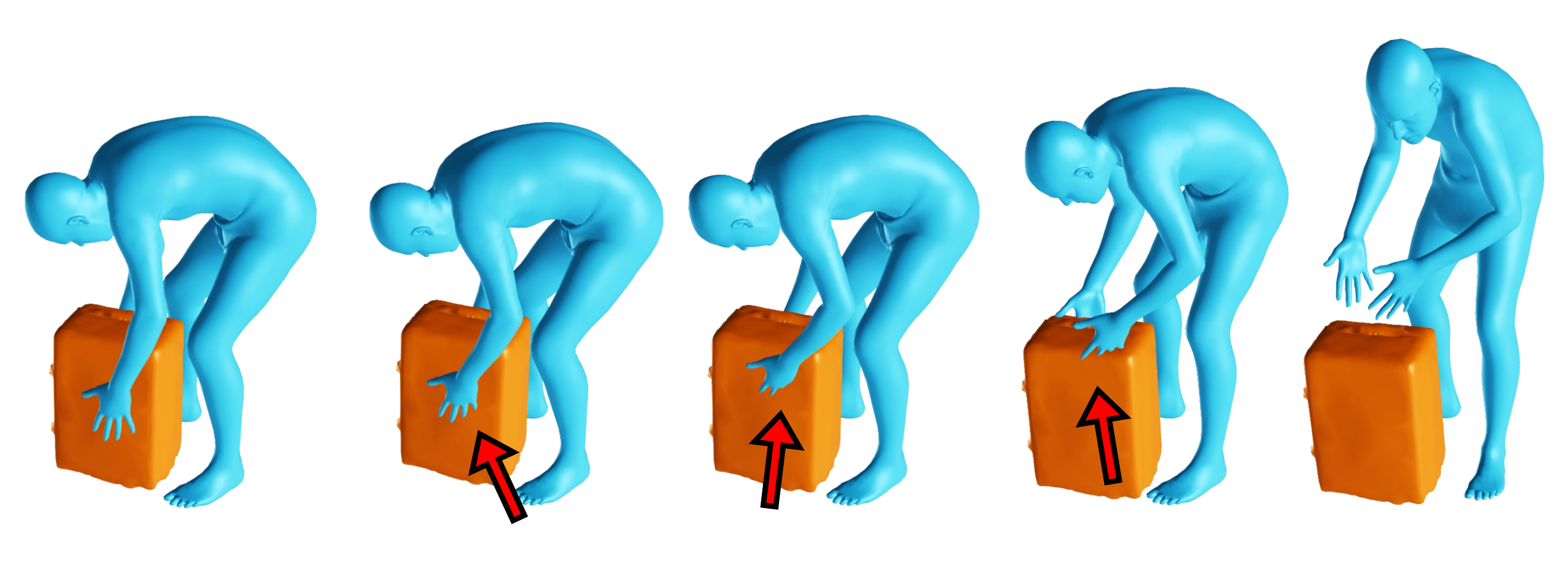}
    \vspace{-6pt}
    \captionof{figure}{\small\textbf{Enforcing kinematic poses in  simulator.}}
    \label{fig:ablation2}
    \vspace{-20pt}

\end{minipage}

\end{figure}
\vspace{-3pt}

\vspace{-3pt}
\subsection{Ablation Study}
\vspace{-6pt}
\label{sec:ablation}

\noindent\textbf{Comparison to original AI Habitat \cite{puig2023habitat}.}
Since we develop \textit{half physics} on the embodied AI platform AI Habitat 3.0, we conduct an ablation study to highlight the differences between the original system and our implementation.
In AI Habitat, humanoids are designed to be purely kinematic. 
Navigation movements are applied on the NavMesh to prevent penetration; however, the motion itself lacks physical interaction with the environment.
As a result, it suffers from penetration issues, similar to other existing HSI environments. As shown in Figure \ref{fig:ablation1}, the agent in AI Habitat moves directly through the wall.
In contrast, our approach maintains physical interactions, preventing mesh penetration.
It is also important to note that in the current AI Habitat, HOI is achieved through hard-coded instructions. 
For example, when a grasping event is detected, the object is instantly repositioned into the hand. 
The experiments in Section \ref{sec:hoi} demonstrate that \textit{half physics} has the potential to enable a more physical HOI implementation on this platform.

\noindent\textbf{Velocity enforcement $\rightarrow$ torque.}
We demonstrate the impact of directly manipulating velocity on the effectiveness of the \textit{half physics} algorithm. 
Specifically, we disable velocity enforcement for each body part, except global velocity, and instead use a PD controller to control each joint.
Table \ref{table:ablation2} presents the tracking performance on the AIST++ dataset. 
When relying solely on joint torques, the global MPJPE-g drastically worsens to $83.10$ \textit{mm}, performing  worse than PHC+. 
Meanwhile, the local MPJPE rises to $82.27$ \textit{mm}.
Notably, since we do not disable root velocity, the agent can still maintain global displacement. 
However, the exceptionally high error in ``HP vel. $\rightarrow$ torque'' almost entirely stems from its inability to track the target pose, demonstrating the crucial role of velocity manipulation for high-fidelity tracking.
For a more intuitive comparison, please refer to the  \href{https://lisiyao21.github.io/projects/Half-Physics/}{\textcolor{magenta}{\texttt{video}}}.



\noindent\textbf{Forcefully use discrete positions in simulation.}
We also investigate what happens if we directly enforce discrete kinematic poses in the physics simulation, instead of approximating linear velocity—an approach akin to hybrid dynamics. Figure \ref{fig:ablation2} illustrates this using a suitcase-lifting example.
Since poses are intractable within continuous timesteps, the simulator fails to respond to collisions correctly, as penetration occurs abruptly (as explained in Figure \ref{fig:hp}(c)).
Although the suitcase shows slight displacement, the interaction remains physically unrealistic.
This highlights the critical importance of converting discrete poses into linear velocity when running kinematic data in a physics simulator, which is the core of our proposed \textit{half physics}.





\vspace{-8pt}
\section{Discussion and Conclusions}
\vspace{-5pt}
\label{sec:conclusion}

In this paper, we propose an approach that extends 3D kinematic human models to a variant that enables tangible physical interactions. 
Unlike existing ``pure'' physics-based methods, we do not manipulate joint torques, which indirectly affect velocity.
Instead, we directly override body velocity, enforcing it to move linearly towards a kinematic target. 
This allows us to efficiently leverage SMPL-X's original pose control while maintaining physical tractability over continuous time intervals, thereby enabling physical interaction.
The proposed \textit{half-physics} strategy can be useful in applications such as VR, where precise alignment with kinematic capture data is required. 
Additionally, unlike methods that drive the body with joint motors, our approach is better able to preserve human-like characteristics in the kinematic motion.
%
\textit{Half physics} serves as a bridge between kinematics and physics, balancing kinematic fidelity and physical realism.
%
See more discussions  in Appendix \ref{sec:moredis}.

\textbf{Acknowledgment.}
%
First, we would like to express our sincere gratitude to several colleagues whose valuable discussions and insights contributed to this work, but who preferred not to be listed as co-authors.
We thank Allan D. Jepson from Toronto University for his insightful conversations during his visit to MPI, especially for raising a critical question: \textit{should a human maintain a predefined kinematic pose after unexpected collision, or should the pose adapt to the physical environment?}
This idea directly inspired the current implementation especially the passive joint stiffness compensation.
We are also grateful to Yan Zhang from Meshcapade, who first suggested the possibility of exploring AI Habitat at the beginning of Siyao’s visit--an idea that eventually evolved into this work.
We thank Mert Albaba for his valuable insights into how fully physical (\textit{i.e.}, RL-based) tracking methods function and what can  benefit from HP.
Second, we acknowledge others for their support and contributions:
Suraj Bhor, for assistance with the AMASS dataset;
Shrisha Bharadwaj, for inspiring the narrative structure and overall storytelling;
Sai Kumar Dwivedi, for sharing relevant papers;
Markos Diomataris, for discussions on HSI and navigation;
Giorgio Becherini, for help with processing the lift-box data;
Artur Grigorev, for discussions on physical simulators;
Nikos Athanasiou, for offering valuable comments that helped improve the work;
Peter Kulits, Prerana Achar, Xianghui Xie and Buzhen Huang, for proofreading;
Anastasios Yiannakidis and Hanz Cuevas Velasquez for introducing the process of converting MoCap points to SMPL-X;
Taylor Obersat, Tsvetelina Alexiadis, Claudia Gallatz, and Markus H\"oschle, for their comments and inspiration on how \textit{half physics} could be used to augment data (\textit{e.g.}, simulating interactions like kicking balls with different weights that are infeasible to collect in MoCap labs).
We also appreciate Zhengyi Luo, Jinkun Cao, Jiaman Li, Zhen Wu, Nan Jiang and  Zhiyuan Zhang for their help on using PHC+, OMOMO, and Trumans.

This research is 
supported by the National Research Foundation, Singapore under
its AI Singapore Programme (AISG Award No: AISG-PhD/2021-01-031[T]).
It also partially benefits from the 2024 Google PhD Fellowship awarded to Siyao, provided as an unrestricted gift.
%


\bibliographystyle{plain}
\bibliography{egbib}

\begin{thebibliography}{10}

\bibitem{araujo2023circle}
Joao~Pedro Araujo, Jiaman Li, Karthik Vetrivel, Rishi Agarwal, Deepak Gopinath, Jiajun Wu, Alexander Clegg, and C.~Karen Liu.
\newblock {CIRCLE}: Capture in rich contextual environments.
\newblock In {\em CVPR}, 2023.

\bibitem{bogo2016keep}
Federica Bogo, Angjoo Kanazawa, Christoph Lassner, Peter Gehler, Javier Romero, and Michael~J Black.
\newblock Keep it smpl: Automatic estimation of 3d human pose and shape from a single image.
\newblock In {\em ECCV}, 2016.

\bibitem{coumans2015bullet}
Erwin Coumans.
\newblock Bullet physics simulation.
\newblock In {\em ACM SIGGRAPH 2015 Courses}, page~1. ACM, 2015.

\bibitem{dallaire2021habitat}
Mikael Dallaire~Côté, Eric Undersander, Alexander Clegg, et~al.
\newblock Habitat 2.0: Training home assistants to rearrange their habitat.
\newblock In {\em NeurIPS}, pages 21525--21537, 2021.

\bibitem{featherstone1983calculation}
Roy Featherstone.
\newblock The calculation of robot dynamics using articulated-body inertias.
\newblock {\em International Journal of Robotics Research}, 2(1):13--30, 1983.

\bibitem{feng2024chatpose}
Yao Feng, Jing Lin, Sai~Kumar Dwivedi, Yu~Sun, Priyanka Patel, and Michael~J Black.
\newblock Chatpose: Chatting about 3d human pose.
\newblock In {\em CVPR}, 2024.

\bibitem{hassan2021stochastic}
Mohamed Hassan, Duygu Ceylan, Ruben Villegas, Jun Saito, Jimei Yang, Yi~Zhou, and Michael~J. Black.
\newblock Stochastic scene-aware motion prediction.
\newblock In {\em ICCV}, 2021.

\bibitem{hassan2019resolving}
Mohamed Hassan, Vasileios Choutas, Dimitrios Tzionas, and Michael~J Black.
\newblock Resolving 3d human pose ambiguities with 3d scene constraints.
\newblock In {\em ICCV}, 2019.

\bibitem{hong2022avatarclip}
Fangzhou Hong, Mingyuan Zhang, Liang Pan, Zhongang Cai, Lei Yang, and Ziwei Liu.
\newblock Avatarclip: Zero-shot text-driven generation and animation of 3d avatars.
\newblock {\em arXiv preprint arXiv:2205.08535}, 2022.

\bibitem{huang2023diffusion}
Siyuan Huang, Zan Wang, Puhao Li, Baoxiong Jia, Tengyu Liu, Yixin Zhu, Wei Liang, and Song-Chun Zhu.
\newblock Diffusion-based generation, optimization, and planning in 3d scenes.
\newblock In {\em ICCV}, 2023.

\bibitem{jakobsen2001advanced}
Thomas Jakobsen.
\newblock Advanced character physics.
\newblock In {\em Game developers conf.}, 2001.

\bibitem{jiang2024autonomous}
Nan Jiang, Zimo He, Zi~Wang, Hongjie Li, Yixin Chen, Siyuan Huang, and Yixin Zhu.
\newblock Autonomous character-scene interaction synthesis from text instruction.
\newblock In {\em SIGGRAPH Asia}, 2024.

\bibitem{jiang2024scaling}
Nan Jiang, Zhiyuan Zhang, Hongjie Li, Xiaoxuan Ma, Zan Wang, Yixin Chen, Tengyu Liu, Yixin Zhu, and Siyuan Huang.
\newblock Scaling up dynamic human-scene interaction modeling.
\newblock In {\em CVPR}, 2024.

\bibitem{kanazawa2018end}
Angjoo Kanazawa, Michael~J Black, David~W Jacobs, and Jitendra Malik.
\newblock End-to-end recovery of human shape and pose.
\newblock In {\em CVPR}, 2018.

\bibitem{kocabas2020vibe}
Muhammed Kocabas, Nikos Athanasiou, and Michael~J Black.
\newblock Vibe: Video inference for human body pose and shape estimation.
\newblock In {\em CVPR}, 2020.

\bibitem{li2024controllable}
Jiaman Li, Alexander Clegg, Roozbeh Mottaghi, Jiajun Wu, Xavier Puig, and C.~Karen Liu.
\newblock Controllable human-object interaction synthesis.
\newblock In {\em ECCV}, 2024.
\newblock Oral Presentation.

\bibitem{li2023object}
Jiaman Li, Jiajun Wu, and C.~Karen Liu.
\newblock Object motion guided human motion synthesis.
\newblock {\em ACM ToG}, 2023.

\bibitem{li2021ai}
Ruilong Li, Shan Yang, David~A Ross, and Angjoo Kanazawa.
\newblock Ai choreographer: Music conditioned 3d dance generation with aist++.
\newblock In {\em ICCV}, 2021.

\bibitem{liu2024revisit}
Xinpeng Liu, Haowen Hou, Yanchao Yang, Yong-Lu Li, and Cewu Lu.
\newblock Revisit human-scene interaction via space occupancy.
\newblock In {\em ECCV}, 2024.

\bibitem{SMPL:2015}
Matthew Loper, Naureen Mahmood, Javier Romero, Gerard Pons-Moll, and Michael~J Black.
\newblock {SMPL}: A skinned multi-person linear model.
\newblock {\em ACM ToG}, 34(6), 2015.

\bibitem{luo2020carl}
Ying-Sheng Luo, Jonathan~Hans Soeseno, Trista Pei-Chun Chen, and Wei-Chao Chen.
\newblock Carl: Controllable agent with reinforcement learning for quadruped locomotion.
\newblock {\em ACM ToG}, 39(4):38--1, 2020.

\bibitem{luo2024omnigrasp}
Zhengyi Luo, Jinkun Cao, Sammy Christen, Alexander Winkler, Kris Kitani, and Weipeng Xu.
\newblock {Omnigrasp}: Grasping diverse objects with simulated humanoids.
\newblock In {\em NeurIPS}, 2024.

\bibitem{luo2023perpetual}
Zhengyi Luo, Jinkun Cao, Kris Kitani, Weipeng Xu, et~al.
\newblock Perpetual humanoid control for real-time simulated avatars.
\newblock In {\em ICCV}, 2023.

\bibitem{luouniversal}
Zhengyi Luo, Jinkun Cao, Josh Merel, Alexander Winkler, Jing Huang, Kris~M Kitani, and Weipeng Xu.
\newblock Universal humanoid motion representations for physics-based control.
\newblock In {\em ICLR}, 2024.

\bibitem{luo2022embodied}
Zhengyi Luo, Shun Iwase, Ye~Yuan, and Kris Kitani.
\newblock Embodied scene-aware human pose estimation.
\newblock {\em NeurIPS}, 2022.

\bibitem{luo2024smplolympics}
Zhengyi Luo, Jiashun Wang, Kangni Liu, Haotian Zhang, Chen Tessler, Jingbo Wang, Ye~Yuan, Jinkun Cao, Zihui Lin, Fengyi Wang, et~al.
\newblock Smplolympics: Sports environments for physically simulated humanoids.
\newblock {\em arXiv preprint arXiv:2407.00187}, 2024.

\bibitem{luo2025humanoidlympics}
Zhengyi Luo, Jiashun Wang, Kangni Liu, Haotian Zhang, Chen Tessler, Jingbo Wang, Ye~Yuan, Jinkun Cao, Zihui Lin, Fengyi Wang, Jessica Hodgins, and Kris Kitani.
\newblock {Humanoidlympics}: Sports environments for physically simulated humanoids.
\newblock In {\em arXiv preprint arXiv:2407.00187}, 2025.

\bibitem{macklin2016xpbd}
Miles Macklin, Matthias M{\"u}ller, and Nuttapong Chentanez.
\newblock Xpbd: position-based simulation of compliant constrained dynamics.
\newblock In {\em ICMG}, 2016.

\bibitem{mahmood2019amass}
Naureen Mahmood, Nima Ghorbani, Nikolaus~F Troje, Gerard Pons-Moll, and Michael~J Black.
\newblock Amass: Archive of motion capture as surface shapes.
\newblock In {\em ICCV}, 2019.

\bibitem{mir2023generating}
Aymen Mir, Xavier Puig, Angjoo Kanazawa, and Gerard Pons-Moll.
\newblock Generating continual human motion in diverse 3d scenes.
\newblock In {\em ICCV}, 2023.

\bibitem{muller2020detailed}
Matthias M{\"u}ller, Miles Macklin, Nuttapong Chentanez, Stefan Jeschke, and Tae-Yong Kim.
\newblock Detailed rigid body simulation with extended position based dynamics.
\newblock In {\em CGF}, 2020.

\bibitem{pan2021synthesizing}
Liang Pan, Jingbo Wang, Buzhen Huang, Junyu Zhang, Haofan Wang, Xu~Tang, and Yangang Wang.
\newblock Synthesizing physically plausible human motions in 3d scenes.
\newblock In {\em 3DV}, 2021.

\bibitem{smplx}
Georgios Pavlakos, Vasileios Choutas, Nima Ghorbani, Timo Bolkart, Ahmed A.~A. Osman, Dimitrios Tzionas, and Michael~J. Black.
\newblock Expressive body capture: {3D} hands, face, and body from a single image.
\newblock In {\em CVPR}, 2019.

\bibitem{peng2018deepmimic}
Xue~Bin Peng, Pieter Abbeel, Sergey Levine, and Michiel Van~de Panne.
\newblock Deepmimic: Example-guided deep reinforcement learning of physics-based character skills.
\newblock {\em ACM ToG}, 37(4):1--14, 2018.

\bibitem{peng2021amp}
Xue~Bin Peng, Ze~Ma, Pieter Abbeel, Sergey Levine, and Angjoo Kanazawa.
\newblock Amp: Adversarial motion priors for stylized physics-based character control.
\newblock {\em ACM ToG}, 40(4):1--20, 2021.

\bibitem{puig2023habitat}
Xavier Puig, Eric Undersander, Andrew Szot, et~al.
\newblock Habitat 3.0: A co-habitat for humans, avatars, and robots.
\newblock {\em arXiv preprint arXiv:2310.13724}, 2023.

\bibitem{ren2022diffmimic}
Jiawei Ren, Cunjun Yu, Siwei Chen, Xiao Ma, Liang Pan, and Ziwei Liu.
\newblock Diffmimic: Efficient motion mimicking with differentiable physics.
\newblock In {\em ICLR}, 2022.

\bibitem{mano}
Javier Romero, Dimitrios Tzionas, and Michael~J Black.
\newblock Embodied hands: Modeling and capturing hands and bodies together.
\newblock {\em arXiv preprint arXiv:2201.02610}, 2022.

\bibitem{saini2022airpose}
Nitin Saini, Elia Bonetto, Eric Price, Aamir Ahmad, and Michael~J Black.
\newblock Airpose: Multi-view fusion network for aerial 3d human pose and shape estimation.
\newblock {\em IEEE Robotics and Automation Letters}, 7(2):4805--4812, 2022.

\bibitem{savva2019habitat}
Manolis Savva, Abhishek Kadian, Oleksandr Maksymets, et~al.
\newblock Habitat: A platform for embodied ai research.
\newblock In {\em ICCV}, 2019.

\bibitem{slerp}
Ken Shoemake.
\newblock Animating rotation with quaternion curves.
\newblock In {\em Annual Conf. on Comput. Graph. and Interactive Techs}, 1985.

\bibitem{siyao2024duolando}
Li~Siyao, Tianpei Gu, Zhitao Yang, Zhengyu Lin, Ziwei Liu, Henghui Ding, Lei Yang, and Chen~Change Loy.
\newblock Duolando: Follower gpt with off-policy reinforcement learning for dance accompaniment.
\newblock In {\em ICLR}, 2024.

\bibitem{siyao2022bailando}
Li~Siyao, Weijiang Yu, Tianpei Gu, Chunze Lin, Quan Wang, Chen Qian, Chen~Change Loy, and Ziwei Liu.
\newblock Bailando: 3d dance generation by actor-critic gpt with choreographic memory.
\newblock In {\em CVPR}, 2022.

\bibitem{siyao2023bailando++}
Li~Siyao, Weijiang Yu, Tianpei Gu, Chunze Lin, Quan Wang, Chen Qian, Chen~Change Loy, and Ziwei Liu.
\newblock Bailando++: 3d dance gpt with choreographic memory.
\newblock {\em IEEE TPAMI}, 45(12):14192--14207, 2023.

\bibitem{GRAB:2020}
Omid Taheri, Nima Ghorbani, Michael~J. Black, and Dimitrios Tzionas.
\newblock {GRAB}: A dataset of whole-body human grasping of objects.
\newblock In {\em ECCV}, 2020.

\bibitem{tevet2025closd}
Guy Tevet, Sigal Raab, Setareh Cohan, Daniele Reda, Zhengyi Luo, Xue~Bin Peng, Amit~H. Bermano, and Michiel van~de Panne.
\newblock {CLoSD}: Closing the loop between simulation and diffusion for multi-task character control.
\newblock In {\em ICLR}, 2025.

\bibitem{wagener2022mocapact}
Nolan Wagener, Andrey Kolobov, Felipe Vieira~Frujeri, Ricky Loynd, Ching-An Cheng, and Matthew Hausknecht.
\newblock Mocapact: A multi-task dataset for simulated humanoid control.
\newblock {\em NeurIPS}, 2022.

\bibitem{wang2021synthesizing}
Jiashun Wang, Huazhe Xu, Jingwei Xu, Sifei Liu, and Xiaolong Wang.
\newblock Synthesizing long-term 3d human motion and interaction in 3d scenes.
\newblock In {\em CVPR}, 2021.

\bibitem{wang2022towards}
Jingbo Wang, Yu~Rong, Jingyuan Liu, Sijie Yan, Dahua Lin, and Bo~Dai.
\newblock Towards diverse and natural scene-aware 3d human motion synthesis.
\newblock In {\em CVPR}, 2022.

\bibitem{wang2021scene}
Jingbo Wang, Sijie Yan, Bo~Dai, and Dahua Lin.
\newblock Scene-aware generative network for human motion synthesis.
\newblock In {\em CVPR}, 2021.

\bibitem{wang2023physhoi}
Yinhuai Wang, Jing Lin, Ailing Zeng, Zhengyi Luo, Jian Zhang, and Lei Zhang.
\newblock {PhysHOI}: Physics-based imitation of dynamic human-object interaction.
\newblock {\em arXiv preprint arXiv:2312.04393}, 2023.

\bibitem{wang2025skillmimic}
Yinhuai Wang, Qihan Zhao, Runyi Yu, Ailing Zeng, Jing Lin, Zhengyi Luo, Hok~Wai Tsui, Jiwen Yu, Xiu Li, Qifeng Chen, Jian Zhang, Lei Zhang, and Ping Tan.
\newblock {SkillMimic}: Learning reusable basketball skills from demonstrations.
\newblock In {\em CVPR}, 2025.

\bibitem{wu2024human}
Zhen Wu, Jiaman Li, Pei Xu, and C.~Karen Liu.
\newblock Human-object interaction from human-level instructions.
\newblock {\em arXiv preprint, arXiv 2406.17840}, 2024.

\bibitem{xiao2024unified}
Zeqi Xiao, Tai Wang, Jingbo Wang, Jinkun Cao, Wenwei Zhang, Bo~Dai, Dahua Lin, and Jiangmiao Pang.
\newblock Unified human-scene interaction via prompted chain-of-contacts.
\newblock In {\em ICLR}, 2024.

\bibitem{xiu2023econ}
Yuliang Xiu, Jinlong Yang, Xu~Cao, Dimitrios Tzionas, and Michael~J Black.
\newblock Econ: Explicit clothed humans optimized via normal integration.
\newblock In {\em CVPR}, 2023.

\bibitem{xiu2022icon}
Yuliang Xiu, Jinlong Yang, Dimitrios Tzionas, and Michael~J Black.
\newblock Icon: Implicit clothed humans obtained from normals.
\newblock In {\em CVPR}, 2022.

\bibitem{xu2023interdiff}
Sirui Xu, Zhengyuan Li, Yu-Xiong Wang, and Liang-Yan Gui.
\newblock Interdiff: Generating 3d human-object interactions with physics-informed diffusion.
\newblock In {\em ICCV}, 2023.

\bibitem{xu2025intermimic}
Sirui Xu, Hung~Yu Ling, Yu-Xiong Wang, and Liangyan Gui.
\newblock Intermimic: Towards universal whole-body control for physics-based human-object interactions.
\newblock In {\em CVPR}, 2025.

\bibitem{yuan2023learning}
Ye~Yuan, Viktor Makoviychuk, Y~Guo, S~Fidler, XB~Peng, and K~Fatahalian.
\newblock Learning physically simulated tennis skills from broadcast videos.
\newblock {\em ACM ToG}, 42(4), 2023.

\bibitem{Zhao:ICCV:2023}
Kaifeng Zhao, Yan Zhang, Shaofei Wang, Thabo Beeler, and Siyu Tang.
\newblock {DIMOS}: Synthesizing diverse human motions in 3d indoor scenes.
\newblock In {\em ICCV}, 2023.

\end{thebibliography}

\appendix

\section{Implementation Details of \textit{Half-Physics} Simulator}
\label{sec:impl_detail}
%

\begin{figure}[t]
   
        \includegraphics[width=0.92\linewidth]{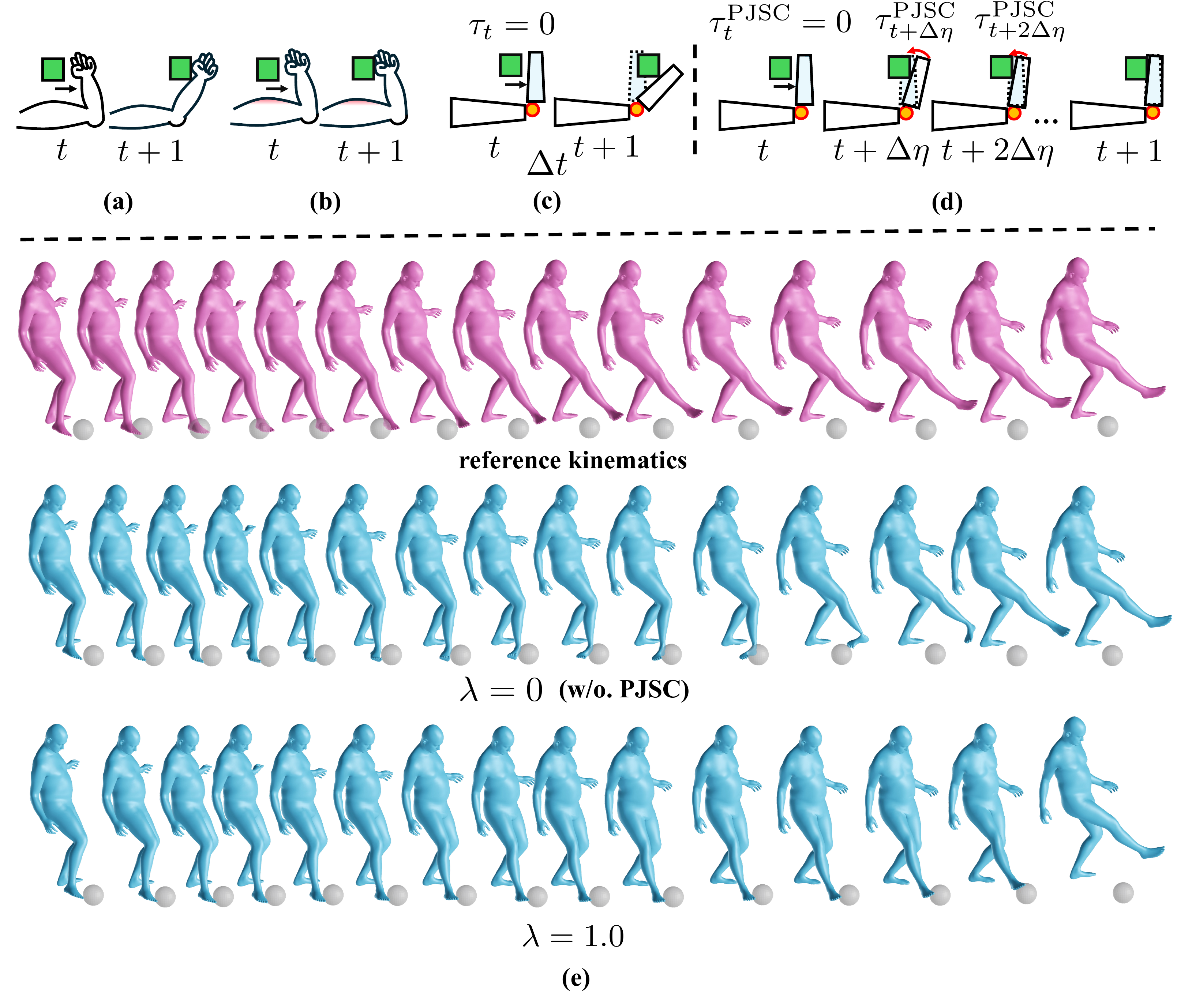}
    \vspace{-8pt}
    \caption{\small\textbf{Passive Joint Stiffness Compensation} (PJSC) in  \textit{half physics}. 
    \textbf{(a, b)} Humans can actively regulate how easily their pose changes in response to external forces by relaxing  or tensing (red) their muscles, even under the same kinematic pose. 
    \textbf{(c)} In contrast, simulated articulated bodies in Featherstone's algorithm lack such adaptive resistance, since when the actual and target poses are identical before contact, the PD controller produces zero torque throughout the timestep $\Delta t$.
    \textbf{(d)} By introducing a dynamic PJSC torque at each substep of duration $\Delta \eta$, the system provides passive compensation that corrects pose deviations after collision.
    \textbf{(e)} Applying PJSC allows the local motion to remain close to the input kinematics, while global translation and orientation adjust to maintain physical interaction. The parameter $\lambda$ controls the magnitude of the PJSC torque.}
    \label{fig:pjsc}

\end{figure}
In theory, \textit{half physics} can be extended to any simulator. 
However, since \textit{half physics} directly manipulates the velocities of each body part, we need the simulator to provide the necessary interfaces. 
Furthermore, if the implementation is based on rigid-body methods such as XPBD \cite{macklin2016xpbd, muller2020detailed}, which operate directly on rigid links rather than joints, forward and inverse kinematics are required before and after simulation to convert between joint-space and link-space velocities.
In contrast, articulation-centric algorithms like Featherstone's method~\cite{featherstone1983calculation} operate directly at the joint level to update all physical states, and therefore inherently integrate both forward and inverse kinematics into the \textcolor{cyan}{\texttt{sim}} pipeline, eliminating the need for additional explicit computation.

We implement the \textit{half physics} algorithm based on AI Habitat~\cite{savva2019habitat,dallaire2021habitat,puig2023habitat},  which internally relies on the Bullet physics engine~\cite{coumans2015bullet} and uses Featherstone's algorithm as its core articulation model.
The following section details our implementation of operator \textcolor{cyan}{\texttt{sim}} in Algorithm \ref{alg:hp}.

\textbf{Damping-Free Dynamics and the Neglect of Internal Force.}
Featherstone’s algorithm typically has two components: one that propagates inter-link interaction forces within the articulated structure (referred to here as ``\textit{internal}'' forces), and another that handles ``\textit{external}'' collision and friction from the environment.
In many scenarios, strong joint damping is introduced during the computation of internal forces.
This causes joint velocities to decay rapidly, making the system more responsive to joint motor torques.
However, in the context of \textit{half physics}, which precisely controls motion by directly specifying joint velocities, such damping will disrupt the intended motion. 
Therefore, we eliminate all joint damping in our implementation.

Furthermore, internal force computation in the Featherstone’s algorithm can also interfere with the intended motion trajectory.
These internal forces model inertial coupling between links, \textit{e.g.}, movement of the forearm passively inducing motion in the upper arm.
However, \textit{half physics} simulates human motion based on kinematic inputs that are assumed to already encode the outcome of complete muscle-skeletal coordination.
Introducing an additional layer of \textit{rigid-body-style} internal forces into this process would distort the intended motion by causing deviations from the original kinematics.
Therefore, we disable internal force propagation in the Featherstone pipeline and retain only the system’s response to external collisions and frictional contacts.

\begin{wraptable}{r}{0.53\textwidth}
  \centering
  \vspace{-1em}

 \centering
   \caption{
     \small{
    \textbf{Ablation study on damping and internal force.}
    }}
      \label{table:ablation3}
    \centering
    \small{
      \centering\scriptsize{
      \begin{tabular}{ l    c c  }
        \toprule
       Method   & MPJPE-\textit{g} (\textit{mm})$\downarrow$ & MPJPE $\downarrow$\\
    \midrule
    HP \textbf{w/.} damp. \textbf{w/.} int. force & $148.802$ & $50.030$\\
    HP \textbf{w/.} int. force &  $145.524$ & $50.848$ \\
     HP  & \ \ \ \ \ $\bf0.003$  &  \ \ \ $\bf 0.003$\\
        \bottomrule
      \end{tabular}
      }
      }

\end{wraptable}

To empirically validate the necessity of removing damping and internal forces, we repeat the experiment from Section~\ref{sec:cmp2pt}.
As shown in Table \ref{table:ablation3}, retaining these components introduces significant errors in the \textit{half-physics} system, which is designed to be velocity-driven. 
Specifically, it results in an average local MPJPE of over $5$ \textit{cm}, and a global error of approximately $15$ \textit{cm} per joint.
Such deviations are fundamentally incompatible with the original intent of \textit{half physics}.
Note that the configuration labeled as ``w/. int. force w/. damp'' corresponds to the default Featherstone implementation in AI Habitat (Bullet).
Therefore, to ensure precise velocity control, these layers must be explicitly disabled in the implementation.


\textbf{Passive Joint Stiffness Compensation}.
%
A fundamental difference between the human body and simulated articulated rigid-body systems in Featherstone's algorithm is that humans can actively modulate their resistance to external disturbances by adjusting muscle tension, whereas simulated systems inherently lack this capability.
As illustrated in Figure~\ref{fig:pjsc}(a, b), when relaxed, a human arm can be easily displaced by external collisions;
in contrast, under muscle tension, it can \textit{actively} resist perturbations and maintain its intended pose—even though the kinematic configuration remains identical to the relaxed case.
Simulated articulated systems, however, do not possess this ability.
As shown in Figure~\ref{fig:pjsc}(c), at time $t$, just before a collision occurs, the target pose (dash blue) matches the current pose (white), causing the PD controller to apply zero torque $\tau_t$ during the simulation interval $\Delta t$.
However, a collision occurs within this timestep, the controller has no opportunity to respond, as the torque remains fixed throughout the entire interval in Featherstone's algorithm.
As a result, the joint pose will inevitably deviate from the intended one after $\Delta t$, unlike a human who can pro-actively resist motion by tightening muscles.
%
%
%
To address this, we propose Passive Joint Stiffness Compensation (PJSC), a strategy designed to approximate resistance to external disturbances by generating restorative torques that “draw back” the pose after a collision as shown in Figure \ref{fig:pjsc}(d).
%
%
%
Specifically, to simulate the pose at time $t+1$, we divide the timestep into smaller substeps with size $\Delta\eta = \Delta t / N$. 
In each substep, we dynamically apply a PJSC torque as follows:
\begin{equation}
    \tau_{t + k\Delta\eta}^{\text{PJSC}, i} = \lambda \left( \hat{q}^{\text{joint}, i}_{t + k\Delta\eta} - \tilde{q}^{\text{joint}, i}_{t + k\Delta\eta} \right),
\end{equation}
where $\lambda$ is the proportional gain coefficient in the PD controller, $\hat{q}^{\text{joint}, i}_{t + k\Delta\eta}$ denotes the actual joint angle before executing the $k$-th substep ($k=0,...,N-1$), and $\tilde{q}^{\text{joint}, i}_{t + k\eta}$ represents the expected angle of the $i$-th joint in the absence of collisions within the interval $[t, t + 1]$:
\begin{equation}
    \tilde{q}^{\text{joint}, i}_{t + k\Delta\eta} = \hat{q}^{\text{joint}, i}_{t} + \dot{q}^{\text{joint}, i}_{t+1} \cdot k\Delta\eta.
\end{equation}
Here, $\hat{q}^{\text{joint}, i}_{t}$ corresponds to the $i$-th joint angle of the output from the previous iteration in $[t-1, t]$, and $\dot{q}^{\text{joint}, i}_{t+1}$ is the joint velocity in the current iteration, which is the joint velocity prescribed for the $[t, t+1]$ simulation to follow the kinematic target (refer to Algorithm~\ref{alg:hp}). 
While the expected pose is expressed here as a linear extrapolation, it is actually implemented using Spherical Linear Interpolation (SLERP) to ensure smooth rotational interpolation at time $t+k\Delta \eta$.
It is important to note that, unlike previous methods where motion is driven by a PD controller, \textit{half physics} is velocity-driven, and the PJSC torque serves only as a corrective signal triggered by collisions.
When no perturbation occurs, \ie, no contact, the expected pose naturally aligns with the current pose, resulting in zero corrective torque.
Upon external collision, however, the deviation activates a restorative torque that passively resists the disturbance, thereby mimicking human-like joint stiffness.

To demonstrate the effect of PJSC, we conduct a comparative experiment to observe how it corrects motion outcomes under collisions.
Specifically, we set $\lambda=0$ and $\lambda=1$ respectively in the kicking experiment described as "fast kick" in Section~\ref{sec:hoi}, and compare how the human motion is corrected with respect to the reference kinematic input.
To eliminate variability caused by the ball's movement, we set it to be static (\ie, approximated as having infinite mass).
As shown in Figure~\ref{fig:pjsc}(e) (note: the ball is rendered semi-transparent to better reveal local motion; please zoom in), the original kinematic trajectory of the right foot passes through the ball.
When  $\lambda=0$, \textit{i.e.}, without PJSC, the motion after \textit{half physics} is corrected such that the right leg bends sharply to brush past the ball, producing a trajectory that globally resembles the reference motion but deviates locally.
In contrast, when $\lambda=1$, the right leg tends to remain straight, preserving the local structure of the kinematic input. 
Instead, the necessary physical interaction is compensated by global changes—such as whole-body displacement or rotation.
This setting echoes a typical challenge in VR/AR applications: when a user moves freely in an open physical space but encounters virtual obstacles in the digital world, how should the avatar respond? 
Should it prioritize maintaining local joint poses, or preserve global displacement and task-level goals?
PJSC provides a controllable trade-off by parameter $\lambda$.

In the experiments shown in Section~\ref{sec:cmp2pt} and Section~\ref{sec:hoi}, we set $\lambda = 0$, \textit{i.e.}, no PJSC is applied. 
In Section~\ref{sec:hsi}, which involves large-scale full-body collisions, we set $\lambda = 1$ and $N = 8$ to enable passive stiffness compensation to keep the stability.

\section{More Discussions}
\label{sec:moredis}
The goal of this paper is to enable kinematic human subjects to physically interact with their environments. In prior work, the modeling of interaction processes has typically followed one of two paradigms: either purely kinematic (also referred to as \textit{zero physics}) or \textit{fully physical}. A typical example of the former is 3D animation, where animators must manually design the full trajectories of both characters and objects; while that of the latter is humanoid simulation, where both the human and the environment are governed by physical dynamics.

To the best of our knowledge, we are the first to propose a \textit{half-physics} paradigm in the human-centric interaction research community, where the human is controlled kinematically while the environment evolves under physical simulation.
In the absence of  contact with the environment, the human motion in \textit{half physics} remains identical to the original kinematic input.
When collisions with the environment occur, the system responds physically, while also allowing for passive modulation of the body's compliance to external contacts by PJSC.

In the setup \textit{half physics}, the human motion enjoys full flexibility — it can follow any predefined kinematic trajectory: running, jumping, even flying like a superhero or performing exaggerated cartoon-like movements. However, unlike a ghost, the human cannot pass through walls. It can fly while holding a ball, but if it does not physically touch the object, the object should not move as if by magic.
Yet, purely kinematic works often suffer from unrealistic interaction outcomes (\eg, penetration) — a long-standing challenge in this line of research. This necessitates the introduction of physics into the interaction process.

A natural question arises: \textbf{Why \textit{half physics}? Why not a fully physical system?} First, as we demonstrate in Experiment \ref{sec:cmp2pt}, even the state-of-the-art physical tracker  fail under complex motions,  resulting in falls. In many tasks involving human motion — such as in embodied AI — researchers are primarily interested in high-level semantic activities and behaviors. \textit{Kinematic control allows them to focus on the motion itself, rather than on the underlying motor control needed to generate it}.

Second, current humanoid simulations rely on motor models that are significantly different from real human biomechanics (\ie, musculoskeletal systems). As a result, motions in fully physical humanoid settings often appear \textit{functional} but lack the \textit{tension}, \textit{expressiveness}, and \textit{subtlety} of real human movements. This lack of expressive detail can significantly undermine realism in applications such as 3D animation or virtual reality. In contrast, kinematic human motions are often directly derived from real-world data (\eg, video or mocap), and thus better preserve the natural characteristics of human movement.

For these reasons, we believe that the proposed \textit{half-physics} paradigm offers a  bridge between the physical realism of environments and the authenticity of human motion.
%

We also emphasize that the core idea of half-physics is conceptually simple: it converts the positional and rotational information encoded in kinematic data into velocity representations with equivalent effect. This simplicity allows half-physics to be easily adapted to different physics simulators.
At the same time, it is important to note that this conversion from position to velocity, while simple, is essential. Without this kinematic-to-physics translation, merely applying trivial techniques—such as increasing the frame rate of the kinematic sequence via interpolation and feeding it directly into a simulator—cannot achieve similar effects. This is because purely kinematic control specifies the exact position (including global position) at every frame, leaving no room for physical responses, which will finally yield penetration (see Section \ref{sec:ablation} ``Forcefully use discrete positions in simulation'').
Moreover, to enable the kind of physical augmentation described in Section~\ref{sec:hoi}, the use of equivalent velocities is essential, as it provides the character with explicit momentum.
If only kinematic inputs are used--even at very high input frame rates--the simulator can, at best, perform penetration correction, but cannot support physically meaningful interactions involving properties such as mass.

\subsection{Discussion on Game, VR, Position-Based Optimizer and Physics Engines}
\label{sec:industry}
\textbf{Game Industry.} In game development, hybrid kinematic and physics control has seen various forms of practice. 
Typically, game engines retain characters in a kinematic (animation-driven) state during active motion, typically disabling physics and relying only on global constraints such as navigation meshes. 
This often results in interpenetration (especially during walking) with the environment. 
Upon impact events, \eg, being killed or falling off cliff, systems switch to a passive ``ragdoll'' mode governed by external physics. 
In contrast, our \textit{half-physics} approach maintains kinematic control while continuously responding to physical interactions.

\textbf{VR Systems.} A similar idea appears in VR games, such as \textit{Half-Life: Alyx}, where hands are kinematically controlled but can move objects. 
However, such forces are usually scripted rather than physically simulated, leading to interpenetration. 
This differs from the physically plausible kinematic-object interactions shown in Section \ref{sec:hoi}.

\textbf{Position-based Optimization.} Position-based optimizer like PBD \cite{jakobsen2001advanced} or XPBD \cite{macklin2016xpbd, muller2020detailed}, unlike our Featherstone-based approach, optimize in position space without computing forces.
However, these methods are fundamentally not intended for active kinematic control, where the pose acts only as a soft constraint within the optimization.

\textbf{Physics Engines.} Physics engines, such as Bullet (used in AI Habitat), typically categorize motion into KINEMATIC, DYNAMIC, and STATIC. 
While limited kinematic-dynamic interaction is allowed, several limitations remain:
\textbf{(1)} KINEMATIC bodies do not collide with STATIC or other KINEMATIC bodies, leading to artifacts such as interpenetration (see Section \ref{sec:ablation} ``Comparison to original AI Habitat''; note AI Habitat is based on Bullet). 
In contrast, \textit{half physics} is velocity-driven and thus capable of responding to environmental collisions.
\textbf{(2)} KINEMATIC bodies are treated as infinitely massive and position-driven, producing identical outcomes regardless of the mass of dynamic objects (\eg, no difference between kicking a 1-ton vs. a 100g ball). 
In contrast, \textit{half physics} enables mass-sensitive interactions (see Section \ref{sec:hoi}: 500g vs. 200kg ball).
\textbf{(3)} KINEMATIC actions are not perturbed by contact, even under strong interactions (\eg, kicking a heavy object) or multi-way contact (\eg, two-hand squeezing), resulting in unrealistic outcomes such as objects being ejected or penetrated.
\textbf{(4)} Forcing dynamic agents to follow kinematic poses causes interpenetration (see Section \ref{sec:ablation} ``Forcefully use discrete positions in simulation'') and constraint drift (due to internal force and other constraints), degrading fidelity.
Our half-physics implementation is lightweight yet avoids these issues, making it well-suited for human-centric interaction tasks requiring active and physically responsive behavior.

\subsection{Potential Applications and Social Impact}
\label{sec:impact}

\begin{figure*}[t]

    \setlength{\tabcolsep}{1.5pt}
    \centering
    \includegraphics[width=0.94\linewidth]{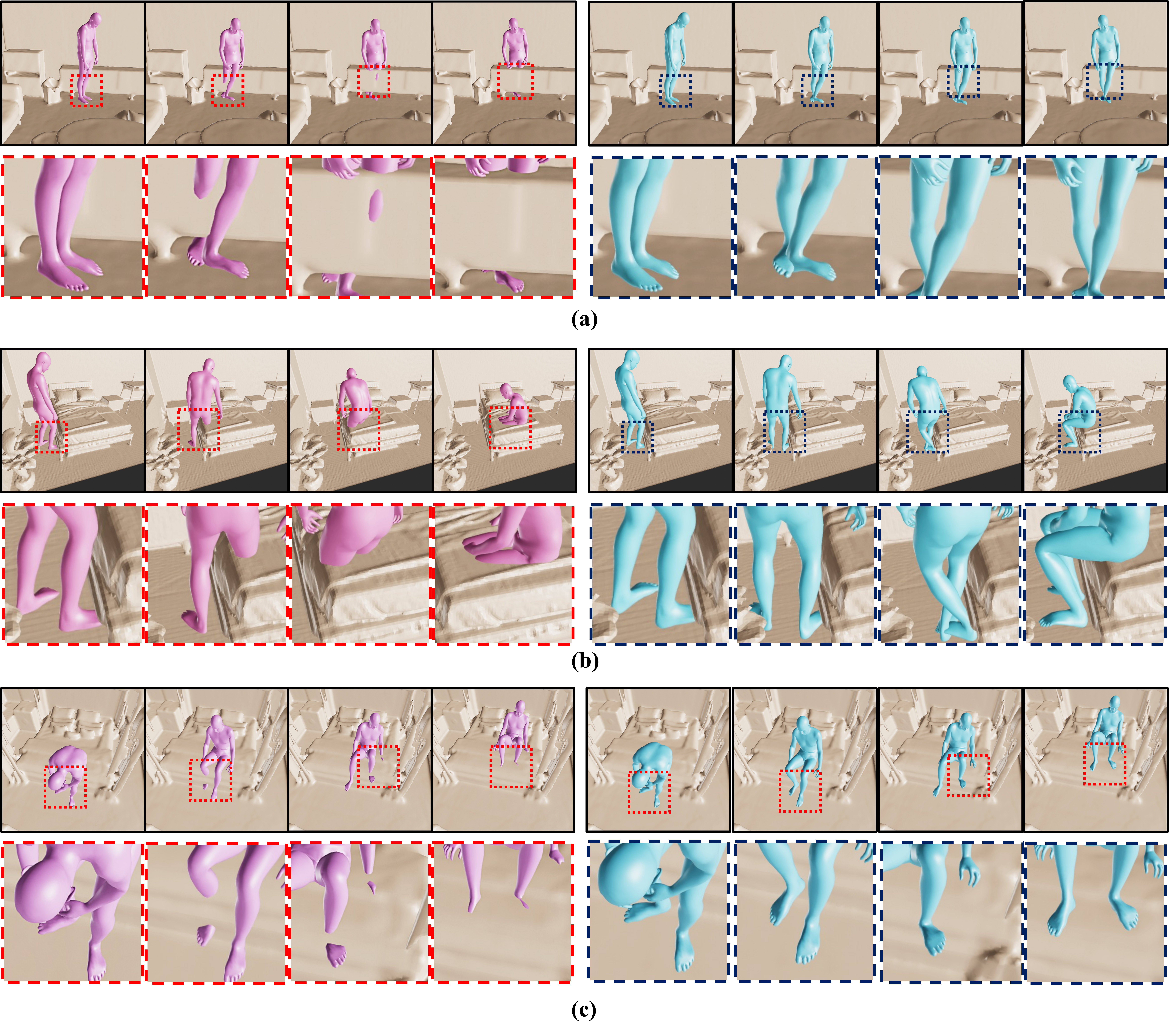}

    \vspace{-15pt}
    \caption{\small{\textbf{
    Qualitative results of Trumans \cite{jiang2024scaling} (pink) and Trumans+HP (ours) (cyan).
    }
    The proposed half physics can effectively eliminate different kinds of penetration in human scene interactions. 
    } }
    \label{fig:trumans_supp}

\end{figure*}

\noindent\textbf{Embodied AI, Human-Scene Interaction and Navigation.}
Embodied AI tasks often involve high-level semantic goals such as indoor navigation or object rearrangement. A key requirement in these scenarios is the avoidance of physical penetrations. However, purely kinematic architectures are fundamentally incapable of preventing such collisions.
To address this, a common practice is to rely on navigation meshes (``navmeshes''), which constrain the human model to predefined ground positions that are collision-free. While this effectively eliminates penetration, it also severely restricts the range of human motion — typically limiting it to walking on ground. As a result, the diversity and realism of embodied AI scenarios are significantly reduced.
Imagine a scene where a 3D human is to crawl under a table to retrieve an apple, or climb onto a bed (as an example shown in Figure \ref{fig:trumans_supp}) to reach a phone — scenarios where navmeshes become ineffective. In contrast, our \textit{half-physics} mechanism can principally avoid interpenetration while enabling such complex and physically grounded interactions, thereby expanding the scope of embodied AI research.

\noindent\textbf{3D Animation.}
Another potential application of half-physics is in 3D animation. In this domain, character motions are typically hand-crafted or captured through mocap, resulting in highly detailed kinematic sequences. However, there is often a strong need to generate physically plausible environmental dynamics in response to these motions. Manually animating dynamic objects is time-consuming and labor-intensive. In contrast, \textit{half physics} offers a promising solution for generating realistic object motions while preserving the original fidelity of the character's performance, as shown in Section \ref{sec:hoi}.

\noindent\textbf{VR and AR.}
In virtual and augmented reality, players’ real-world motions need to be mapped into virtual environments. Regardless of whether the scene is mocap-based or ego-centric, the reconstructed human motion is typically kinematic. In human-scene or human-object interactions within VR, it is crucial to maintain consistency with the user’s real movements while enabling meaningful interactions with virtual objects — such as kicking a ball.
Fully physical (\ie, RL-based) approaches often lead to discrepancies from the player’s intended actions or result in instability (\eg, falling), making them unsuitable for precise VR control. In contrast, half-physics provides an  alternative that preserves motion fidelity while enabling physically grounded interactions in virtual environments.

\noindent\textbf{Social Impact.}
All of the aforementioned examples demonstrate the potential for positive social impact by advancing research or enabling practical applications in their respective domains. As for potential negative impacts, one concern might be the increased risk of user addiction from more immersive VR gaming experiences.

\subsection{Limitations}
\label{sec:limit}
Since \textit{half physics} employs kinematic control, it does not model the internal dynamical mechanisms  responsible for actuating the agent. 
Therefore, it is not capable of research on robotics, as it cannot serve as a true simulation of robotic systems. 
Notably, current humanoid robots still rely heavily on joint motors, making \textit{full-physics} simulation essential for robotics research.
Besides, our method does not address self-penetration in SMPL-X if it is already present in the input motion sequence. 
Moreover, \textit{half physics} does not semantically correct the motion. If a wall is introduced in front of the character but the reference human motion keeps walking forward as if the wall weren't there, this inconsistency can result in foot sliding.
Similar to prior works, the current implementation still works on \textit{\textbf{rigid}}-body collisions and frictions, without modeling soft-body or deformable interactions.
This limitation may lead to discrepancies between simulation and reality in tasks that require high-precision contact handling, such as dexterous hand manipulation.
We leave these as a direction for future improvement.


\section{Additional Human-Scene Interaction (HSI) Results}
In addition to the examples presented in the main paper, we provide three additional, more complex cases in the appendix (see Figure \ref{fig:trumans_supp}).
In Figure \ref{fig:trumans_supp}(a), we demonstrate a scenario where a human attempts to step backward and lean against a TV cabinet. In the original Trumans results, the character penetrates through the cabinet and into the wall while stepping backward. In contrast, with half physics simulation, our method (Trumans+HP) successfully prevents penetration, allowing the character to lean against the cabinet as intended.
In Figure \ref{fig:trumans_supp}(b), the human attempts to sit on a bed and slide inward. However, in the original results, the legs penetrate deep into the bed, causing severe artifacts. With half physics, the character correctly maintains a natural seated posture at the edge of the bed.
In Figure \ref{fig:trumans_supp}(c), the Trumans-generated character initially sits on the bed without major penetration issues. However, when extending the arms and legs inward, partial penetration occurs in both the hands and feet. Although such minor penetration artifacts are difficult to eliminate solely through loss optimization, they are highly perceptible and significantly impact realism. In contrast, by leveraging a physics engine, half physics effectively resolves even these small penetration issues, demonstrating its effectiveness in handling penetration in human-scene interactions.
%





\end{document}